%% file: main.tex
\pdfoutput=1 

\documentclass[10pt,twocolumn,letterpaper]{article}

\usepackage{cvpr}
\usepackage{times}
\usepackage{epsfig}
\usepackage{graphicx}
\usepackage{amsmath}
\usepackage{amssymb}
\usepackage{physics,paralist}
\usepackage{physics,paralist}
\usepackage{multirow}
\usepackage{wrapfig,amsthm}
\usepackage[pagebackref=true,breaklinks=true,letterpaper=true,colorlinks,bookmarks=false]{hyperref}
\usepackage{xcolor,colortbl}
\usepackage{stfloats}

\cvprfinalcopy 
\def\cvprPaperID{3051}

\ifcvprfinal\fi

\definecolor{Gray}{gray}{0.85}
\definecolor{LightCyan2}{rgb}{0.94,0.81,0.81}
\definecolor{LightCyan}{rgb}{0.84,0.84,0.84}
\DeclareMathOperator{\E}{\mathbb{E}}

\newcolumntype{a}{>{\columncolor{Gray}}c}
\newcolumntype{b}{>{\columncolor{white}}c}
\newtheorem{lemma}{Lemma}

\ifcvprfinal\fi

\makeatletter
\newcommand*\bigcdot{\mathpalette\bigcdot@{.5}}
\newcommand*\bigcdot@[2]{\mathbin{\vcenter{\hbox{\scalebox{#2}{$\m@th#1\bullet$}}}}}
\makeatother

\begin{document}

\title{Orthogonal Convolutional Neural Networks}

\author{
\setlength{\tabcolsep}{10pt}
\begin{tabular}{@{}cccc@{}}
Jiayun Wang &  
Yubei Chen &
Rudrasis Chakraborty & 
Stella X. Yu\\
\end{tabular}\\[5pt]
UC Berkeley / ICSI\\
\tt\small\{peterwg,yubeic,rudra,stellayu\}@berkeley.edu\\
}

\maketitle
\input{sections/abstract.tex}
\vspace{-0.2in}

\input{sections/1intro.tex}
\input{sections/2relatedwork.tex}

\input{sections/3approach.tex}
\input{sections/4experiments.tex}

\input{sections/5conclusion.tex}

{\small
\bibliographystyle{ieee}
\bibliography{egbib, sup_bib}
}

\clearpage
\input{append.tex}

\end{document}


\title{Orthogonal Convolutional Neural Networks\\Supplementary Material}

\author{First Author\\
Institution1\\
Institution1 address\\
{\tt\small firstauthor@i1.org}
\and
Second Author\\
Institution2\\
First line of institution2 address\\
{\tt\small secondauthor@i2.org}
}

\maketitle
The supplementary material provides new experimental results showing the power of our orthogonal convolutional neural networks in improving the convergence and performance of generative adversarial networks (GANs), as well as enhancing robustness under attack, along with additional insights  and technical details  to the main paper.

We provide intuitive explanations of our methods in Section \ref{sec:exp} and relations to feature redundancy to complement related work in Section \ref{sec:red}.  We demonstrate the effectiveness of our model in image generation in Section \ref{sec:gen} and the robustness of our approach under adversarial example attack in Section \ref{sec:attack}. We finally show more image retrieval results in Section \ref{sec:ret}.

\section{Intuitive Explanations of our Approach}
\label{sec:exp}

We  analyze a convolution layer which transforms input $X$ to output $Y$ with learnable kernel $K$: $Y = \text{Conv}(K, X)$ in CNNs. Writing in linear matrix multiplication form $Y = \mathcal{K} X$ (Fig.1(b) of the paper), we simplify the analysis from the perspective of linear systems.  We do not use \textit{im2col} form $Y =K \widetilde{X}$ (Fig.1(a) of the paper) as there is an additional structured linear transform from $X$ to $\tilde{X}$, which does not necessarily have a uniform spectrum.

The spectrum of $\mathcal{K}$ reflects the scaling property of the convolution layer: different input $X$ (such as cat, dog, and house images) would scale up by $\eta = \frac{\|Y\|}{\|X\|}$. The scaling factor $\eta$ also reflects the gradient scaling. Typical CNNs have very imbalanced convolution spectrum (Fig.2(b) of the paper): for some inputs, it scales up to 2; for others, it scales by $0.1$. For a deep network, these irregular spectrums add up and can potentially lead to gradient exploding and vanishing issues.

Features learned by CNNs are also more redundant due to the imbalanced spectrum (Fig.2(a) of the paper). This comes from the diverse learning ability to different images and leads to feature redundancy. A uniform spectrum distribution could alleviate the problem.

To alleviate the problem, we propose to make convolution orthogonal by making $\mathcal{K}$ orthogonal. Orthogonal convolution regularizer in CNNs (OCNNs) leads to uniform $\mathcal{K}$ spectrum as expected, and further reduce the feature redundancy and improve the performance (Fig.2(b)(c)(d) of the paper).

Besides classification performance improvements, we also observe improved visual features, both in high-level (image retrieval) and low-level (image inpainting). Our OCNNs also generate realistic images (Section \ref{sec:gen}) and is robust to attacks (Section \ref{sec:attack}). 

\section{Relations to Feature Redundancy}
\label{sec:red}

CNNs are shown to have significant redundancy between different filters and feature channels \cite{jaderberg2014speeding, howard2017mobilenets}. Many works use the redundancy to compress or speed up networks \cite{han2016eie, he2017channel, howard2017mobilenets}. The imbalanced distribution of spectrum may contribute to redundancy in CNNs.

There is an increasing popularity in improving the feature diversity to overcome the problem, including multi-attention \cite{zheng2017learning}, diversity loss \cite{li2018diversity}, and orthogonality \cite{chen2017training}. We adopt an orthogonal regularizer, which penalizes the kernel $\mathcal{K}$ so that different filters are orthogonal and share minimal similarity. The analysis and empirical experimental results indicate that our model reduces the feature redundancy and improves the feature diversity.

\section{Image Generation}

\label{sec:gen}

\begin{figure}[!t]
        \includegraphics[width=0.25\textwidth]{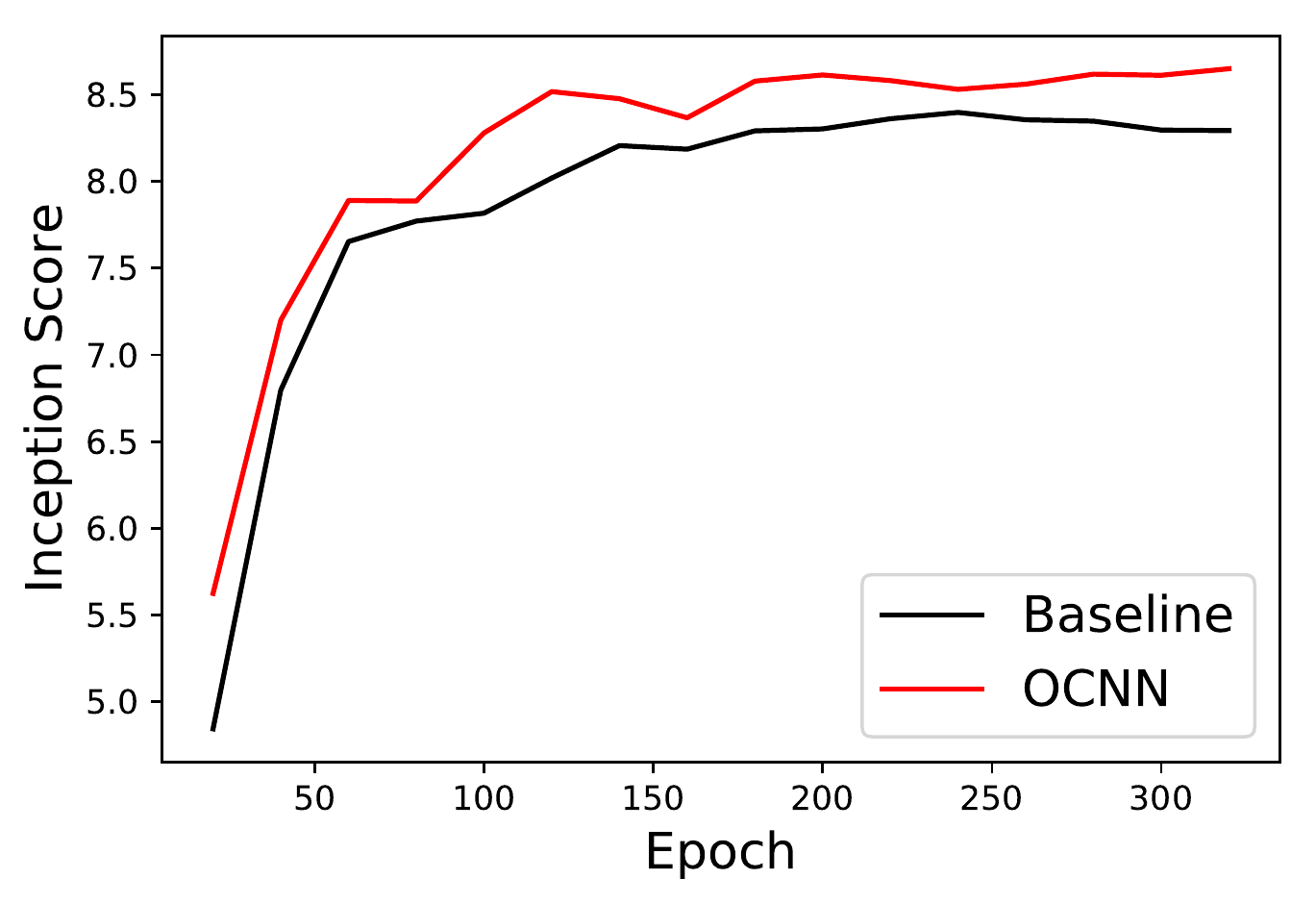}%
        \includegraphics[width=0.25\textwidth]{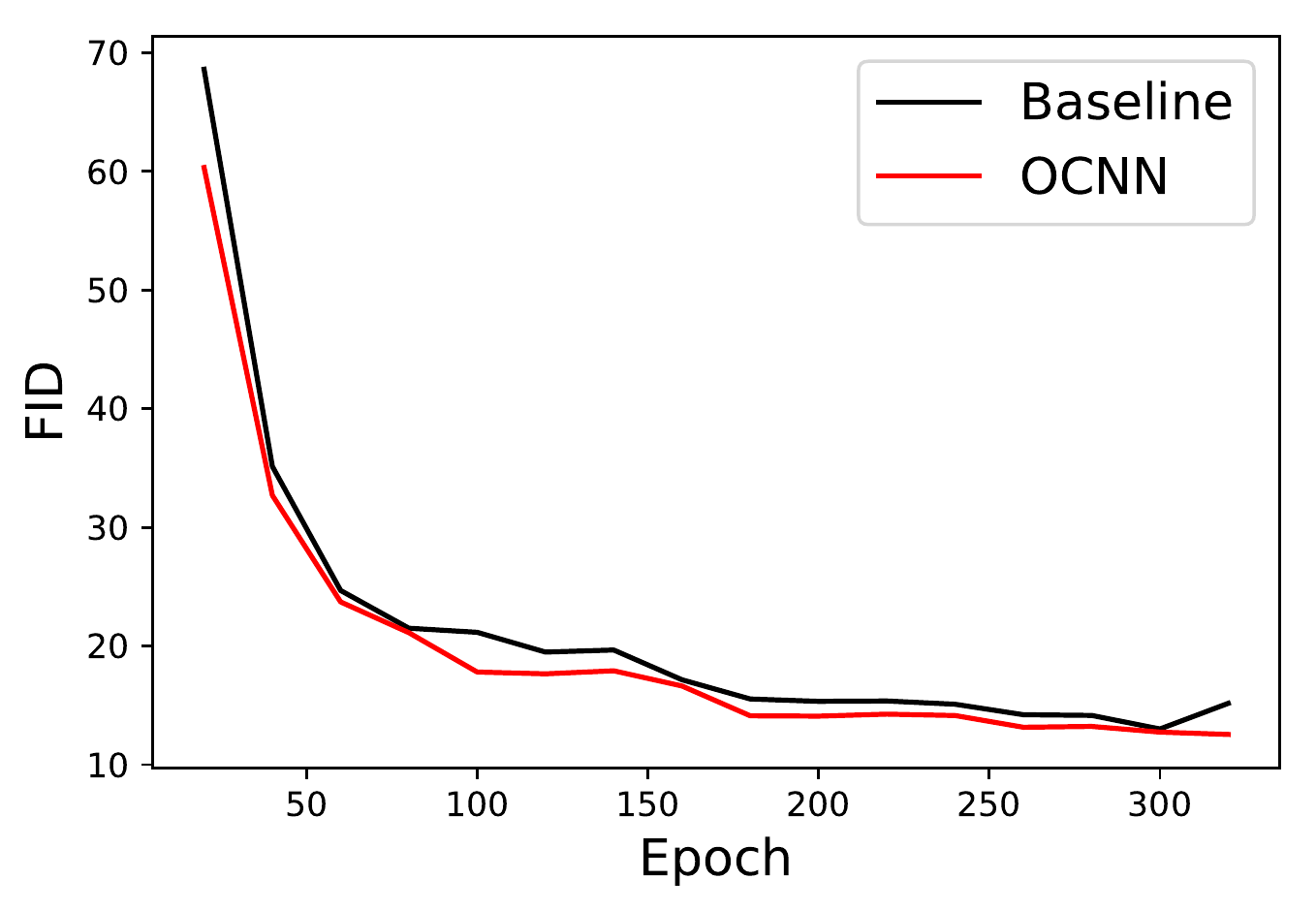}%
       \caption{\small  OCNNs have faster convergence for GANs. The IS (left) and FID (right) consistently outperforms the baseline \cite{gong2019autogan} at every epoch.}
\label{fig:fastgan}
   \end{figure}   

\begin{table}[H]
\begin{center}
\centering
\caption{\small Inception Score and Fréchet Inception Distance comparison on CIFAR10. Our OCNN outperforms the baseline \cite{gong2019autogan} by 0.3 IS and 1.3 FID.}
\label{tab:gen}

\resizebox{0.3\textwidth}{!}{
\begin{tabular}{>{\columncolor[gray]{0.95}}c|c|c}
\hline
\rowcolor{LightCyan}       & IS   & FID   \\
PixelCNN \cite{van2016conditional} & 4.60 & 65.93 \\
PixelIQN \cite{pmlr-v80-ostrovski18a} & 5.29 & 49.46 \\
EBM \cite{du2019implicit}     & 6.78 & 38.20 \\
SNGAN \cite{miyato2018spectral}   & 8.22 & 21.70 \\
BigGAN \cite{brock2018large}  & \textbf{9.22} & 14.73 \\
AutoGAN \cite{gong2019autogan} & 8.32 & 13.01 \\ \hline
OCNN (ours)     & 8.63 & \textbf{11.75}  \\ \hline
\end{tabular}
}

\end{center}
\end{table}

Orthogonal regularizers have shown great success in improving the stability and performance  of GANs \cite{brock2016neural, miyato2018spectral, brock2018large}. We analyze if our convolutional orthogonal regularizer can help train GANs.

We use the best architecture reported in \cite{gong2019autogan}, and train for 320 epochs with our convolutional orthogonal regularizer applying on both the generator and discriminator with regularizer weight 0.01, and retain all other settings. 

The reported model is evaluated 5 times with 50k images each.
We achieve $8.63 \pm 0.007$ inception score (IS) and $11.75 \pm 0.04$  Fréchet inception distance (FID) (Table \ref{tab:gen}),  outperforming the baseline and achieving the state-of-the-art performance. Additionally, we observe faster convergence of GANs with our regularizer (Fig.\ref{fig:fastgan}).
   
\section{Robustness under Attack}
\label{sec:attack}

The spectrum of $\mathcal{K}$ is uniform, so each convolution layer approximates $1$-Lipschitz function. Given a perturbation of input $\Delta x$, the change of output $\Delta y$ is bounded to be low. Therefore, the model enjoys robustness under attack because it is hard to search for adversarial examples and the change in output is small and slow.
   
\begin{table}[!ht]
\begin{center}
\centering
\caption{\small Attack time and number of necessary attack queries needed for 90\% successful attack rate.}
\label{tab:attack}

\resizebox{0.45\textwidth}{!}{
\begin{tabular}{>{\columncolor[gray]{0.95}}c|c|c}
\hline
\rowcolor{LightCyan}       &  Attack time/s & \# necessary attack queries  \\ \hline
ResNet18 \cite{he2016deep}   &      19.3        &  27k          \\ \hline
OCNN (ours) &    136.7         &    46k        \\ \hline
\end{tabular}
}
\end{center}
\end{table}

\begin{figure}[t!]
        \includegraphics[width=0.25\textwidth]{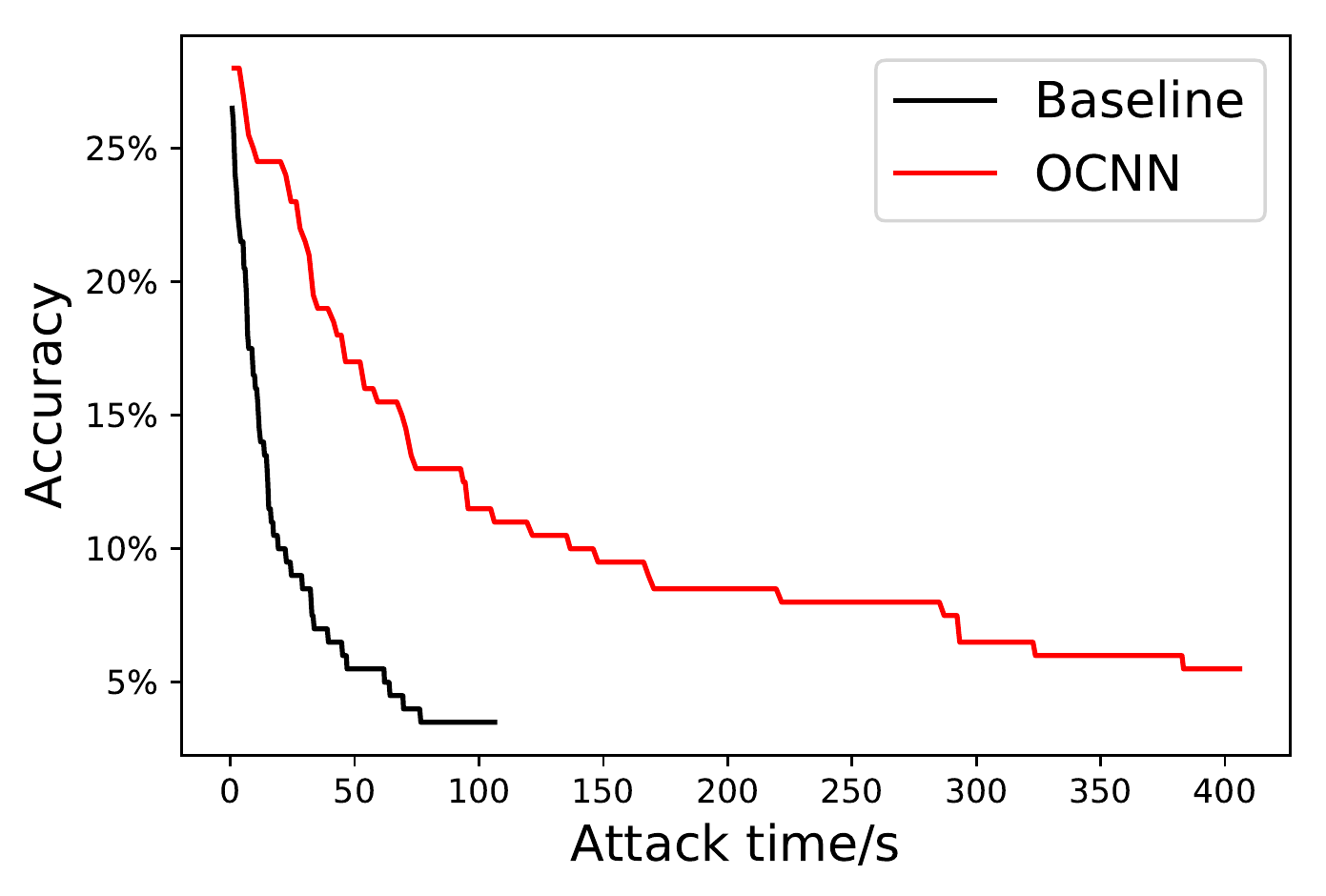}%
        \includegraphics[width=0.25\textwidth]{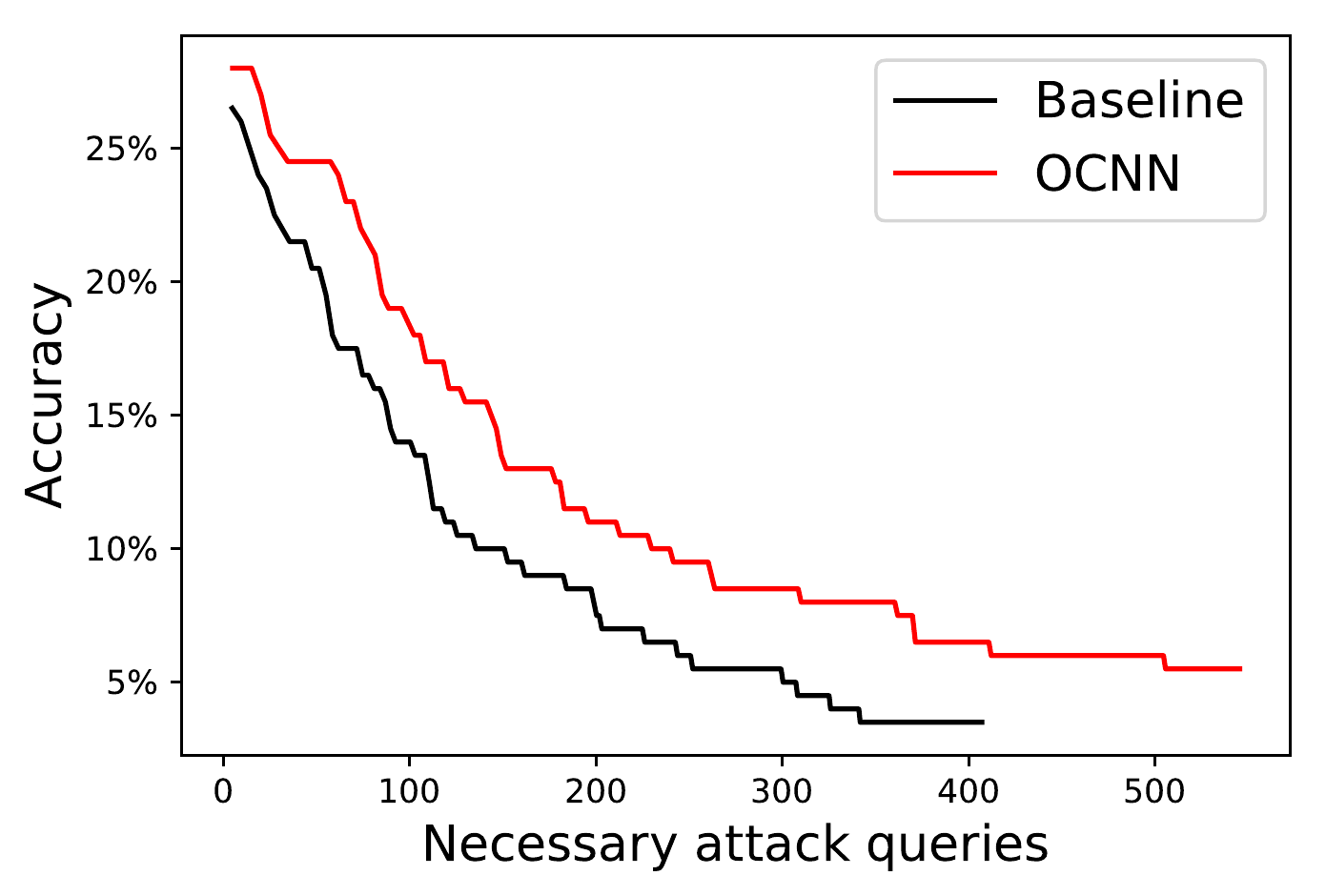}%
       \caption{\small  Model accuracy v.s. attack time and necessary attack queries. With our conv-orthogonal regularizer, it takes  7x time and 1.7x necessary attack queries to achieve 90\% successful attack rate. Note that baseline ends at accuracy 3.5\% while ours ends at 5.5\% with the same iteration.}
\label{fig:attack}
   \end{figure}
We adopt the simple black box attack \cite{guo2019simple} to evaluate the robustness of baseline and our OCNN with ResNet18 \cite{he2016deep} backbone architecture trained on CIFAR100. The attack basically samples around the input image and finds the ``direction'' to rapidly decrease the classification confidence of the network by manipulating the input. We only evaluate on the correctly classified test images. The maximum iteration is 10000 with pixel attack. All other settings are retained. We report the attack time and number of necessary attack queries for a specific attack successful rate.

It takes about 7x time and 1.7x attack queries to attack our OCNN, compared with the baseline (Fig.\ref{fig:attack} and Table \ref{tab:attack}). Additionally, after the same iterations of the attack, our model has 2\% more accuracy than the baseline. 

To achieve the same attack rate, baseline models need more necessary attack queries, and searching for such queries is nontrivial and may take time. This may account for the longer attack time of the OCNN.

\section{More Image Retrieval Results}
\label{sec:ret}
\begin{figure*}[t!]
        \includegraphics[width=1\textwidth]{figures/sup_ret.pdf}%
       \caption{\small Additional image retrieval results on CUB-200 Birds.  The model (ResNet34) is trained on ImageNet only without further fine-tuning. Every two rows show ours (1st) and baseline (2nd) results for the same query. Ours gains 2\% and 3\% more in the top-1 and top-5 $k$-nearest neighbor classification accuracy.}
\label{fig:rett}
   \end{figure*}
   
    We show more image retrieval results on CUB-200 birds dataset \cite{WelinderEtal2010} of our model trained on ImageNet \cite{deng2009imagenet} only (Fig. \ref{fig:rett}). Compared with the baseline, our OCNN learn  more expressive features and retrieve more similar and yet diverse images, including fine-grained bird variations, bird poses, surroundings, etc. 
{\small
\bibliographystyle{ieee}
\bibliography{sup_bib, egbib}
}

%% file: sections/abstract.tex
\begin{abstract}

Deep convolutional neural networks are hindered by training instability and feature redundancy towards further performance improvement.  A promising solution is to impose orthogonality on convolutional filters. 

We develop an efficient approach to impose filter orthogonality on a convolutional layer based on the doubly block-Toeplitz matrix representation of the convolutional kernel instead of using the common kernel orthogonality approach, which we show is only necessary but not sufficient for ensuring orthogonal convolutions.

Our proposed orthogonal convolution requires no additional parameters and little computational overhead.  This method consistently outperforms the kernel orthogonality alternative on a wide range of tasks such as image classification and inpainting under supervised, semi-supervised and unsupervised settings.  Further, it learns more diverse and expressive features with better training stability, robustness, and generalization. 
Our \href{https://github.com/samaonline/Orthogonal-Convolutional-Neural-Networks}{code} is publicly available.


\end{abstract}

%% file: sections/1intro.tex

\def\summaryTable#1{
\begin{table}[#1]
\centering
\caption{\small Summary of experiments and OCNN gains.}
\label{tab:exp_sum}
\resizebox{\columnwidth}{!}{
\begin{tabular}{c|c|c|c}
\hline
\rowcolor{LightCyan} \multicolumn{2}{|c|}{Task}                                                                                      & Metric                      & Gain    \\ \hline
\multirow{3}{*}{\begin{tabular}[c]{@{}c@{}} Image\\ Classification\end{tabular}} & CIFAR100                      & classification accuracy     & 3\%     \\ \cline{2-4} 
                                                                                & ImageNet                      & classification accuracy     & 1\%     \\ \cline{2-4} 
                                                                                & semi-supervised learning      & classification accuracy     & 3\%     \\ \hline 
\multirow{4}{*}{\begin{tabular}[c]{@{}c@{}}Feature\\ Quality\end{tabular}}      & fine-grained image retrieval  & kNN classification accuracy & 3\%     \\ \cline{2-4} 
                                                                                & unsupervised image inpainting & PSNR                        & 4.3     \\ \cline{2-4} 
                                                                                & image generation              & FID                         & 1.3     \\ \cline{2-4} 
                                                                                & Cars196                       & NMI                         & 1.2     \\ \hline
Robustness               & black box attack              & attack time                 & 7x less \\ \hline
\end{tabular}
}
\vspace{-0.15in}
\end{table}
}

\def\figoverview#1{
\begin{figure}[#1]
\centering

        \includegraphics[width=\columnwidth]{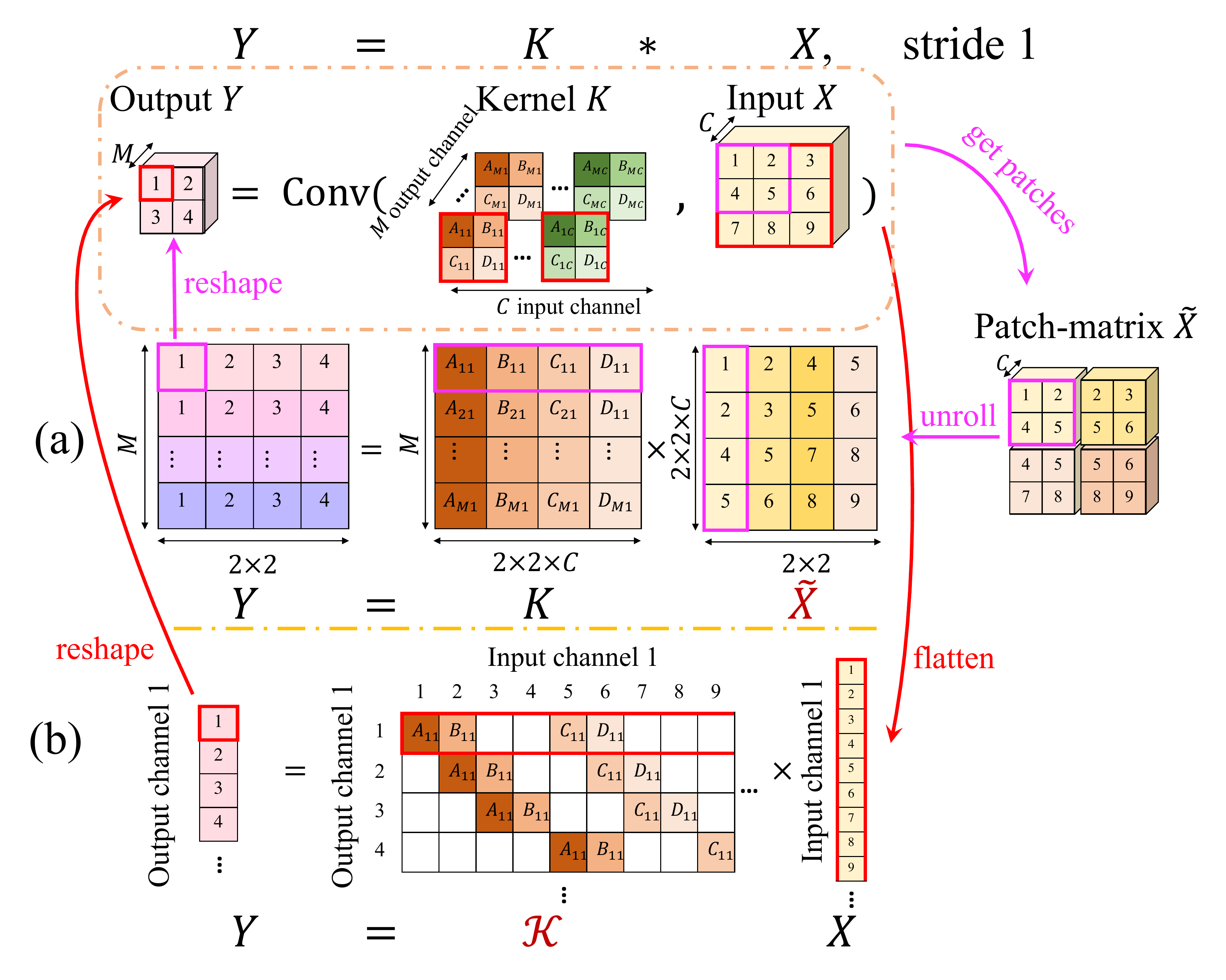}%
       \caption{\small  Basic idea of our OCNN.   A convolutional layer $Y = \text{Conv}(K, X)$ can be formulated as matrix multiplications in two ways: {\bf a)} \textit{im2col} methods \cite{yanai2016efficient, heide2015fast}  retain kernel $K$ and convert input $X$ to patch-matrix $\widetilde{X}$. {\bf b)} We retain input $X$ and convert $K$ to a doubly block-Toeplitz matrix $\mathcal{K}$. With $X$ and $Y$ intact, we directly analyze the transformation from the input to the output.  We further propose an efficient algorithm for regularizing $\mathcal{K}$ towards orthogonal convolutions and observe improved feature expressiveness, task performance and uniformity in  $\mathcal{K}$'s spectrum (Fig.\ref{fig:intro}b). 
   \label{fig:overview}
   }
   \vspace{-0.25in}
   \end{figure}
}

\def\figintro#1{
\begin{figure}[#1]
\centering\small
\begin{tabular}{@{}l@{\hspace{5pt}}r@{}}
\includegraphics[height=0.1\textheight,width=0.24 \textwidth]{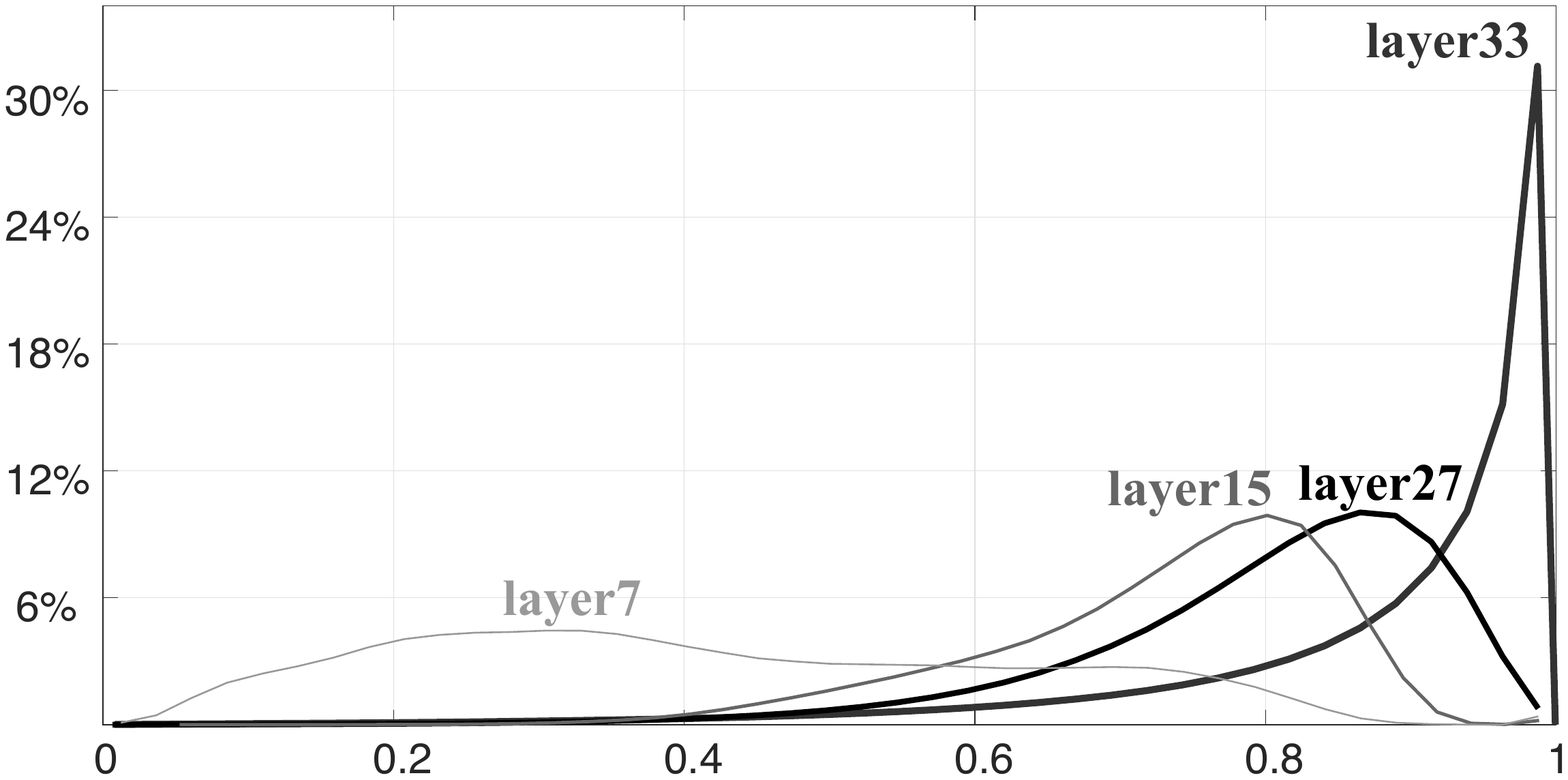} & 
\includegraphics[height=0.1\textheight,width=0.235 \textwidth]{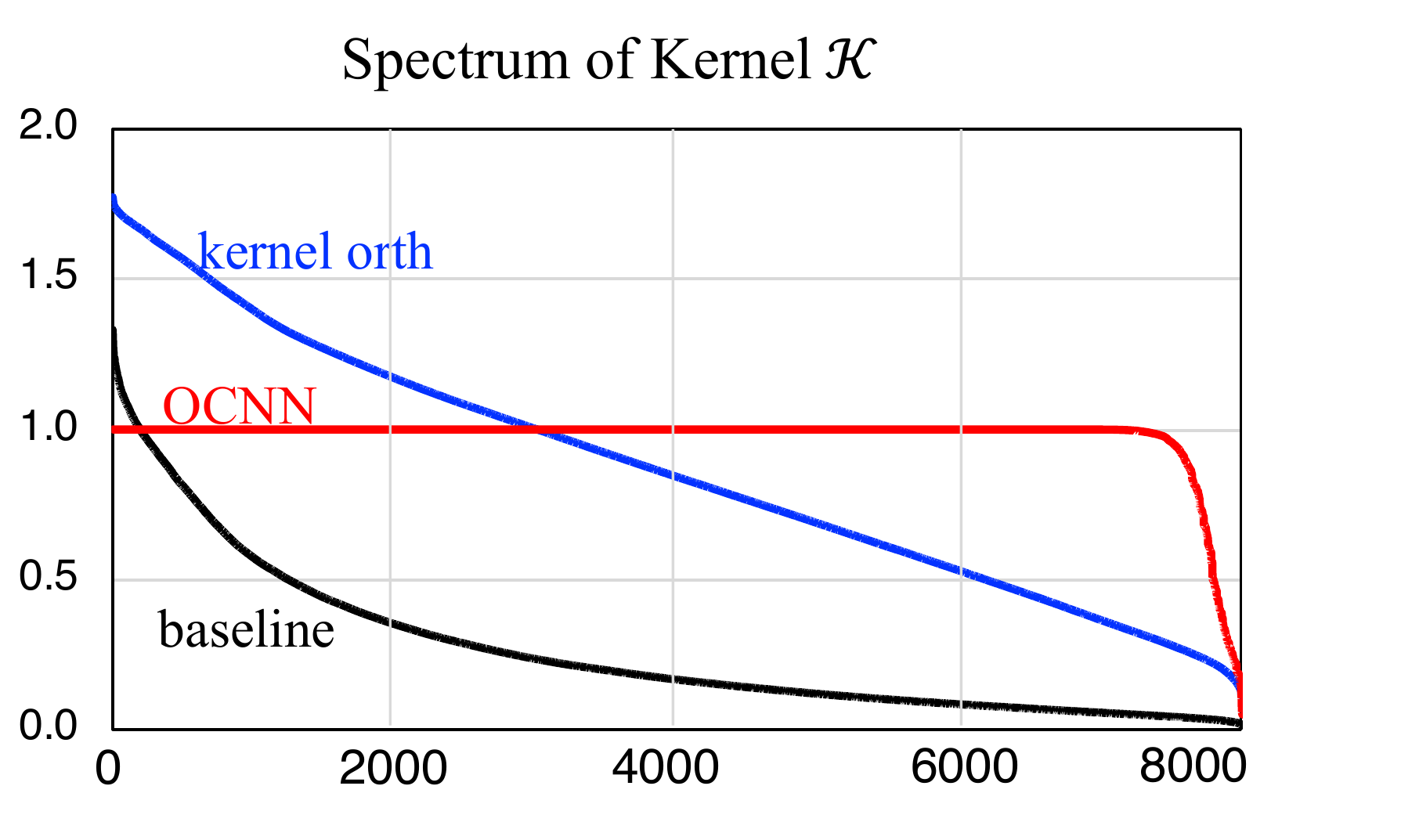} \\ 
{\bf a)} histogram of filter similarities&
{\bf b)} convolution kernel spectrum\\
\end{tabular}\\[3pt]
\begin{tabular}{@{}l@{\hspace{1pt}}r@{}}
\includegraphics[height=0.1\textheight,width=0.24 \textwidth]{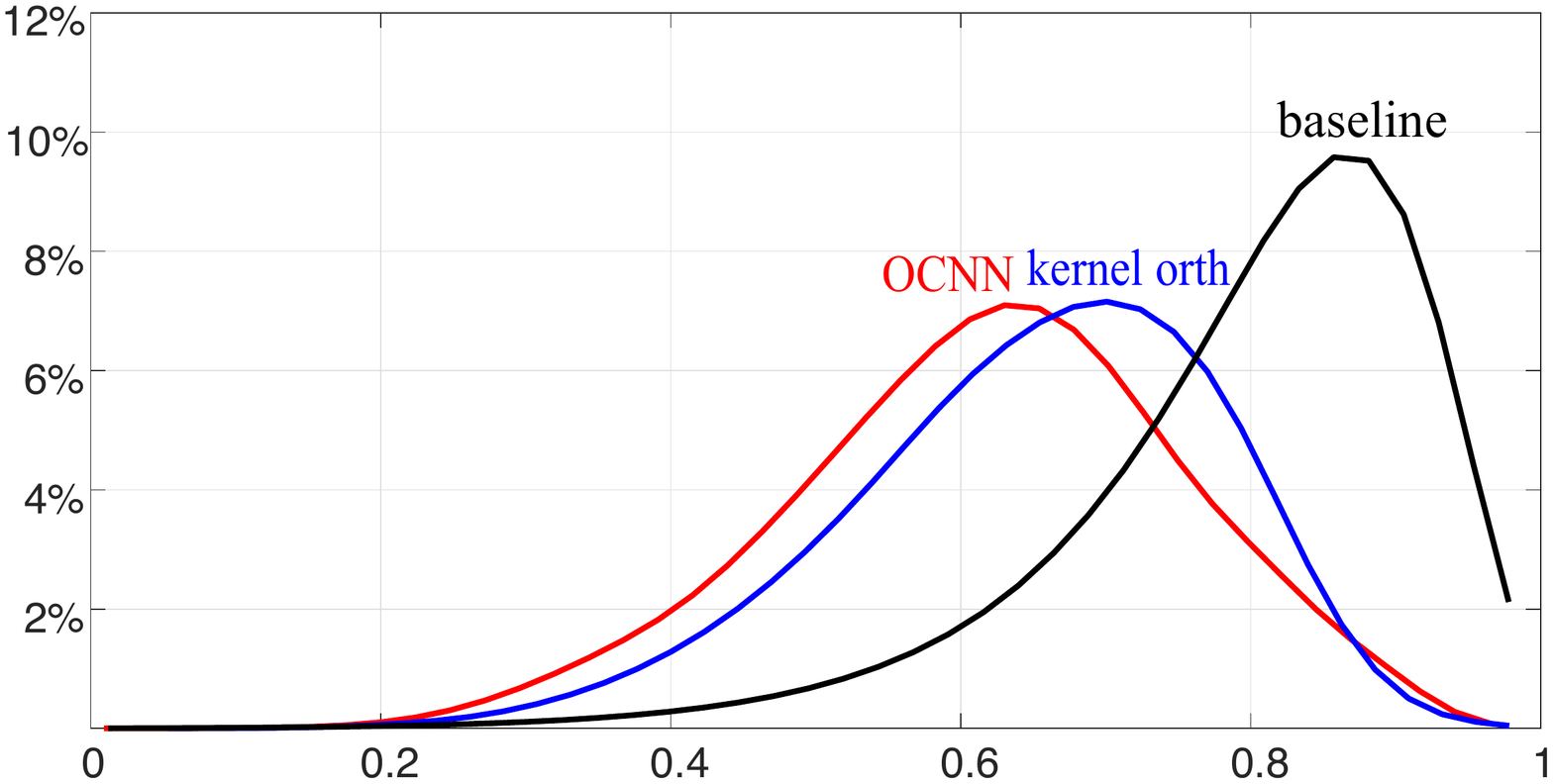} &  
\includegraphics[height=0.1\textheight,width=0.24 \textwidth]{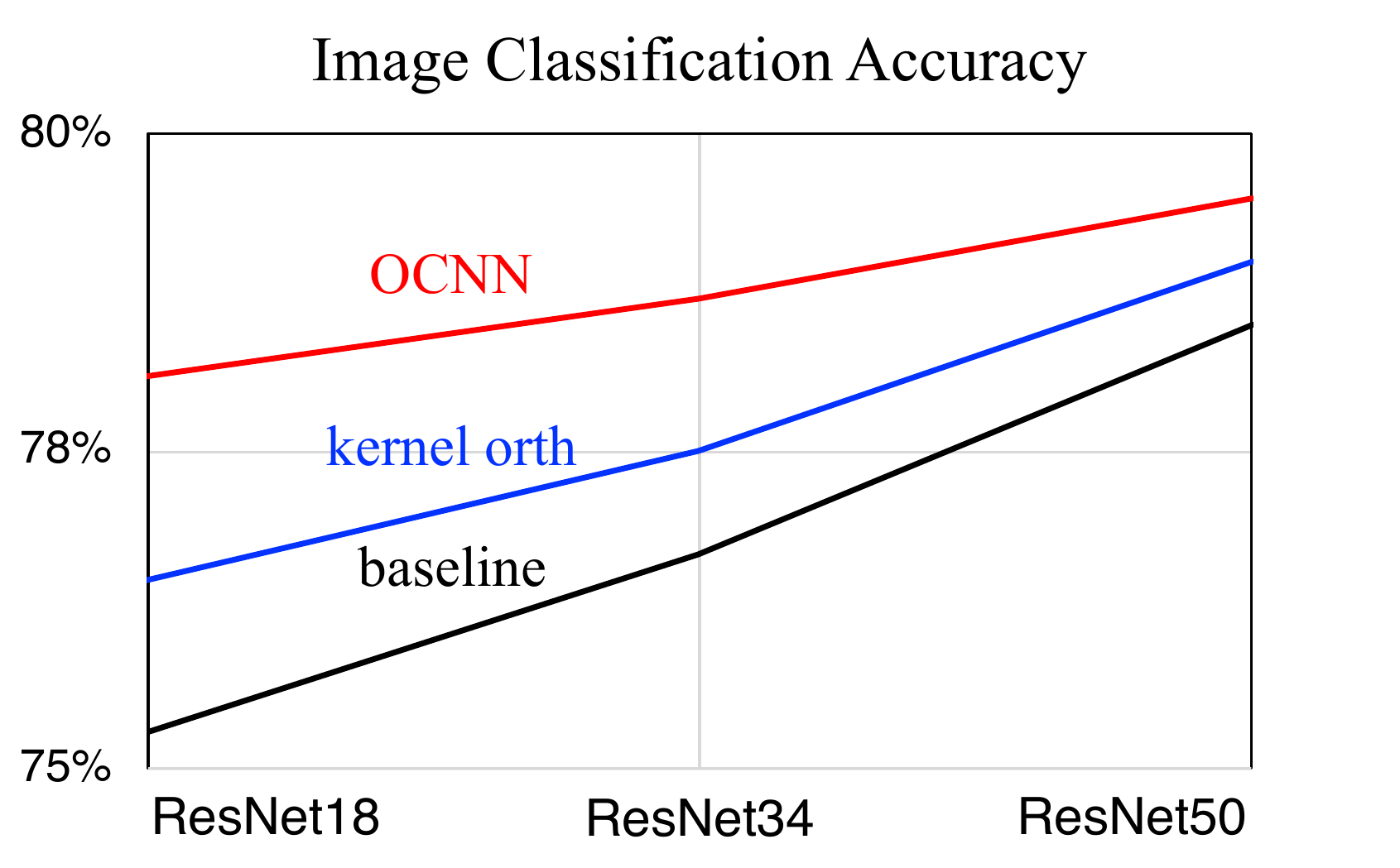}\\
{\bf c)} histogram of filter similarities&
{\bf d)} classification accuracy\\[6pt]
\end{tabular}
   \caption{\small Our OCNN can remove correlations among filters and result in consistent performance gain over standard convolution {\it baseline} and alternative  kernel orthogonality baseline ({\it\blue kernel orth}) during testing.
   {\bf a)} Normalized histograms of pairwise filter similarities of ResNet34 for ImageNet classification  show increasing correlation among standard convolutional filters with depth.   {\bf b)} A standard convolutional layer has a long-tailed spectrum.  While kernel orthogonality  widens the spectrum, our OCNN can produce a more ideal uniform spectrum.  {\bf c)} Filter similarity (for layer 27 in {\bf a}) is reduced most with our OCNN.  {\bf d)} Classification accuracy on CIFAR100 always increases the most with our OCNN. }
  \label{fig:intro}
   
   \end{figure}
}

\section{Introduction}


\begin{figure}[t!]
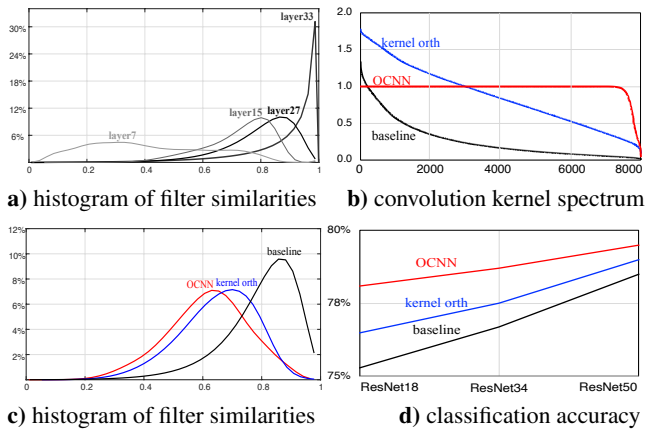

\centering\small
\begin{tabular}{@{}l@{\hspace{5pt}}r@{}}
\includegraphics[height=0.1\textheight,width=0.24 \textwidth]{figures/intro_a.pdf} & 
\includegraphics[height=0.1\textheight,width=0.235 \textwidth]{figures/intro_b.pdf} \\ 
{\bf a)} histogram of filter similarities&
{\bf b)} convolution kernel spectrum\\
\end{tabular}\\[3pt]
\begin{tabular}{@{}l@{\hspace{1pt}}r@{}}
\includegraphics[height=0.1\textheight,width=0.24 \textwidth]{figures/intro_c.pdf} &  
\includegraphics[height=0.1\textheight,width=0.24 \textwidth]{figures/intro_d.pdf}\\
{\bf c)} histogram of filter similarities&
{\bf d)} classification accuracy\\[6pt]
\end{tabular}
   \caption{\small Our \textcolor{red}{\it OCNN} can remove correlations among filters and result in consistent performance gain over standard convolution {\it baseline} and alternative  kernel orthogonality baseline ({\it \textcolor{blue}{kernel orth}}) during testing.
   {\bf a)} Normalized histograms of pairwise filter similarities of ResNet34 for ImageNet classification  show increasing correlation among standard convolutional filters with depth. 
   {\bf b)} A standard convolutional layer has a long-tailed spectrum.  While kernel orthogonality  widens the spectrum, our OCNN can produce a more ideal uniform spectrum.  {\bf c)} Filter similarity (for layer 27 in {\bf a}) is reduced most with our OCNN. 
   {\textbf d)} Classification accuracy on CIFAR100 always increases the most with our OCNN.}
   \label{fig:intro}
   
   \end{figure}
   
While convolutional neural networks (CNNs) are widely successful \cite{krizhevsky2012imagenet, deng2009imagenet, simonyan2014very}, several challenges still exist: over parameterization or under utilization of model capacity \cite{han2016deep, cheung2019superposition},  exploding or vanishing gradients \cite{bengio1994learning, glorot2010understanding}, growth in saddle points \cite{dauphin2014identifying}, and shifts in feature statistics  \cite{ioffe2015batch}. Through our analysis to solve these issues, we observe that convolutional filters learned in deeper layers are not only highly correlated and thus redundant (Fig.\ref{fig:intro}a), but that each layer also has a long-tailed spectrum as a linear operator (Fig.\ref{fig:intro}b), contributing to unstable training performance from exploding or vanishing gradients.


We propose {\it orthogonal CNN} (OCNN), where a convolutional layer is regularized with orthogonality constraints during training.  When filters are learned to be as orthogonal as possible, they become de-correlated. Their filter responses are much less redundant. Therefore, the model capacity is better utilized, which improves the feature expressiveness and consequently the task performance. 

Specifically, we show that simply by regularizing convolutions with our orthogonality loss during training, networks produce more uniform spectra (Fig.\ref{fig:intro}b) and more diverse features (Fig.\ref{fig:intro}c), delivering consistent performance gains with various network architectures (Fig.\ref{fig:intro}d) on various tasks, e.g. image classification/retrieval, image inpainting, image generation, and adversarial attacks (Table \ref{tab:exp_sum}).

\summaryTable{t}

Many works have proposed the orthogonality of linear operations as a type of regularization in training deep neural networks.   Such a regularization improves the stability and performance of CNNs \cite{bansal2018can, xie2017all, balestriero2018mad, balestriero2018spline}, since it can preserve energy, make spectra uniform \cite{zhou2006special}, stabilize the activation distribution in different network layers \cite{rodriguez2016regularizing}, and remedy the exploding or vanishing gradient issues \cite{arjovsky2016unitary}. 

\figoverview{tp}

Existing works impose orthogonality constraints as kernel orthogonality, whereas ours directly implements orthogonal convolutions, based on an entirely different formulation of a convolutional layer as a linear operator.  

Orthogonality for a convolutional layer $Y = \text{Conv}(K, X)$ can be introduced in two different forms (Fig.\ref{fig:overview}).
\begin{enumerate}
\setlength{\itemsep}{0pt}
\item{\bf Kernel orthogonality} methods \cite{xie2017all, balestriero2018mad, balestriero2018spline} view convolution as  multiplication between the kernel matrix $K$ and the \textit{im2col} \cite{yanai2016efficient, heide2015fast} matrix $\widetilde{X}$, \ie $Y = K\widetilde{X}$.  The orthogonality is enforced by penalizing the disparity between the Gram matrix of kernel $K$ and the identity matrix, \ie $\| KK^T-I\|$.  However, the construction of $\widetilde{X}$ from input $X$ is also a linear operation $\widetilde{X} = QX$, and $Q$ has a highly nonuniform spectrum. \item{\bf Orthogonal convolution} keeps the input $X$ and the output $Y$ intact by connecting them with a doubly block-Toeplitz (DBT) matrix $\mathcal{K}$ of filter $K$, i.e. $Y = \mathcal{K}X$ and enforces the orthogonality of $\mathcal{K}$ directly. 
We can thus directly analyze the linear transformation properties between the input $X$ and the output $Y$.
\end{enumerate} 
Existing works on CNNs adopt kernel orthogonality, due to its direct filter representation. 

We prove that kernel orthogonality is in fact only necessary but not sufficient for orthogonal convolutions.  Consequently, the spectrum of a convolutional layer is still non-uniform and exhibits a wide variation even when the kernel matrix $K$ itself is orthogonal (Fig.\ref{fig:intro}b).  

More recent works propose to improve the kernel orthogonality by normalizing spectral norms \cite{miyato2018spectral}, regularizing mutual coherence \cite{bansal2018can}, and penalizing off-diagonal elements \cite{brock2018large}.  Despite the improved stability and performance, the orthogonality of $K$ is insufficient to make a linear convolutional layer orthogonal among its filters. In contrast, we adopt the DBT matrix form,  and regularize $\| \text{Conv}(K, K)-I_r\|$ instead.  While the kernel $K$ is indirectly represented in the DBT matrix $\mathcal{K}$, the representation of input $X$ and output $Y$ is intact and thus the orthogonality property of their transformation can be directly enforced.

We show that our regularization enforces orthogonal convolutions more effectively than kernel orthogonality methods, and we further develop an efficient approach for our OCNN regularization.

To summarize, we make the following contributions.
\begin{enumerate}
\setlength{\itemsep}{0pt}
\item We provide an equivalence condition for orthogonal convolutions and develop efficient algorithms to implement orthogonal convolutions for CNNs. 
\item With no additional parameters and little computational overhead, our OCNN consistently outperforms other orthogonal regularizers on image classification, generation, retrieval, and inpainting under supervised, semi-supervised, and unsupervised settings. 
\end{enumerate}
Better feature expressiveness, reduced feature correlation, more uniform spectrum, and enhanced adversarial robustness may underlie our performance gain. 

%% file: sections/2relatedwork.tex
\section{Related Works}

\noindent \textbf{\textit{Im2col}-Based Convolutions}.  The \textit{im2col} method \cite{yanai2016efficient, heide2015fast} has been widely used in deep learning as it enables efficient GPU computation. It transforms the convolution into a General Matrix to Matrix Multiplication (GEMM) problem.


Fig.\ref{fig:overview}a illustrates the procedure. \begin{inparaenum}[\bfseries a)] \item Given an input $X$, we first construct a new input-patch-matrix $\widetilde{X} \in \mathbf{R}^{Ck^2 \times H'W' }$ by copying patches from the input and unrolling them into columns of this intermediate matrix. \item The kernel-patch-matrix $K \in \mathbf{R}^{M \times Ck^2  }$ can then be constructed by reshaping the original kernel tensor.  Here we use the same notation for simplicity. \item We can calculate the output $Y = K \widetilde{X}$ where we reshape $Y$ 
back to the tensor of size ${M \times H\times W}$ -- the desired output of the convolution.
\end{inparaenum}

The orthogonal kernel regularization enforces the kernel ${K} \in \mathbf{R}^{M \times Ck^2}$ to be orthogonal. Specifically, if $M \leq Ck^2$, the row orthogonal regularizer is $L_{\text{korth-row}} = \| KK^T - I\|_F$
where $I$ is the identity matrix.  Otherwise, column orthogonal may be achieved by $L_{\text{korth-col}}  = \| K^TK - I\|_F$.


\vspace{5pt}

\noindent \textbf{Kernel Orthogonality in Neural Networks.} Orthogonal kernels help alleviate gradient vanishing or exploding problems in recurrent neural networks (RNNs) \cite{dorobantu2016dizzyrnn, wisdom2016full, lezcano2019cheap, arjovsky2016unitary, vorontsov2017orthogonality, pascanu2013difficulty}.  
The effect of soft versus hard orthogonal constraints on the performance of RNNs is discussed in \cite{vorontsov2017orthogonality}. A cheap orthogonal constraint based on a parameterization from exponential maps is proposed in \cite{lezcano2019cheap}.

Orthogonal kernels are also shown to stabilize the training of CNNs \cite{rodriguez2016regularizing} and make more efficient optimizations \cite{bansal2018can}.  Orthogonal weight initialization is proposed in \cite{saxe2013exact, mishkin2015all}; utilizing the norm-preserving property of orthogonal matrices, it is similar to the effect of batch normalization \cite{ioffe2015batch}.  However, the orthogonality may not sustain as the training proceeds \cite{saxe2013exact}. 
 To ensure the orthogonality through the whole training, Stiefel manifold-based optimization methods are used in \cite{harandi2016generalized, ozay2016optimization, huang2018orthogonal} and are further extended to convolutional layers in
  \cite{ozay2016optimization}.
  
Recent works relax and extend the exact orthogonal weights in CNNs. Xie et al.  enforce the Gram matrix of the weight matrix to be close to identity under Frobenius norm \cite{xie2017all} . 
Bansal  et al. further utilize mutual coherence and the restricted isometry property \cite{bansal2018can}.  Orthogonal regularization has also been observed to help improve the performance of image generation in generative adversarial networks (GANs) \cite{brock2018large, brock2016neural, miyato2018spectral}.

All the aforementioned works adopt kernel orthogonality for convolutions.  Sedghi et al.  utilize the DBT matrix to analyze singular values of convolutional layers but do not consider orthogonality \cite{sedghi2018singular}. 

\vspace{5pt}

\noindent\textbf{Feature Redundancy.}  Optimized CNNs are known to have significant redundancy between different filters and feature channels \cite{jaderberg2014speeding, howard2017mobilenets}.  Many works use the redundancy to compress or speed up networks \cite{han2016eie, he2017channel, howard2017mobilenets}.  The highly nonuniform spectra may contribute to the redundancy in CNNs. To overcome the redundancy by improving feature diversity,  multi-attention \cite{zheng2017learning}, diversity loss \cite{li2018diversity}, and orthogonality regularization \cite{chen2017training} have been proposed. 

\vspace{5pt}

\noindent \textbf{Other Ways to Stabilize CNN Training.}  To address  unstable gradient and co-variate shift problems, various methods have been proposed:
Initialize each layer with near-constant variances  \cite{glorot2010understanding,he2015delving};
Use batch normalization to reduce internal covariate shifts
 \cite{ioffe2015batch}; Reparameterize the weight vectors and decouple their lengths from their directions  \cite{salimans2016weight}; Use layer normalization with the mean and variance computed from all of the summed inputs to the neurons \cite{ba2016layer}; Use a gradient norm clipping strategy to deal with exploding gradients and a soft constraint for vanishing gradients \cite{pascanu2013difficulty}.

%% file: sections/3approach.tex
\section{Orthogonal Convolution}
\label{sec:orthogonalconv}
As we mentioned earlier, convolution can be viewed as an efficient matrix-vector multiplication, where matrix $\mathcal{K}$ is generated by a kernel $K$. In order to stabilize the spectrum of $\mathcal{K}$, we add  convolutional orthogonality regularization to CNNs, which is a stronger condition than kernel orthogonality. First, we discuss the view of convolution as a matrix-vector multiplication in detail. Then, fast algorithms for constraining row and column orthogonality in convolutions are proposed. Condition \ref{eq:roworthogonal_conv2} summarizes the orthogonality. In this work, we focus on the 2D convolution case, but concepts and conditions generalize to other cases.

\subsection{Convolution as a Matrix-Vector Multiplication}
\begin{figure}[t!]
   \vspace{-0.5em}

\centering
        \includegraphics[width=\columnwidth]{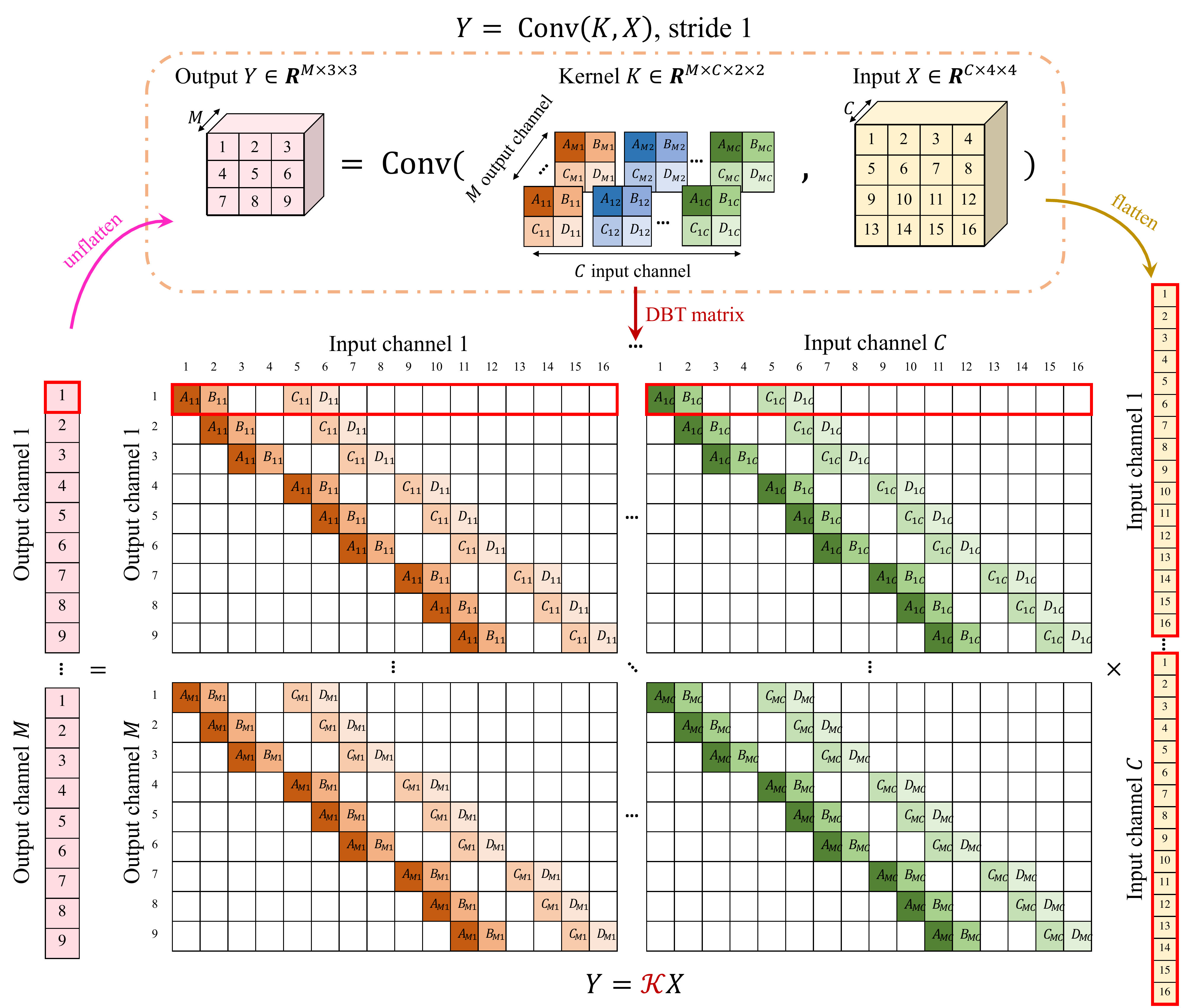}%
   \caption{\small Convolution based on the doubly block-Toeplitz (DBT) matrix. We first flatten $X$ to a vector $\mathbf{x}$, and then convert weight tensor $K \in \mathbf{R}^{M \times C \times k \times k}$ as DBT matrix $\mathcal{K} \in \mathbf{R}^{ (M H' W')\times (C H W)}$. The output $\mathbf{y} = \mathcal{K} \mathbf{x}$. We can obtain the desired output $Y \in \mathbf{R}^{ M \times H' \times W'}$ by reshaping $\mathbf{y} $. The example has input size $C\times4\times4$, kernel size $M\times C\times 2 \times2$ and stride 1.}
   \vspace{-1em}
   \label{fig:cconv}
   \end{figure}


For a convolutional layer with input tensor $X \in \mathbf{R}^{C \times H \times W}$ and kernel $K \in \mathbf{R}^{M\times C \times k \times k}$, we denote the convolution's output tensor $Y = \text{Conv}(K,X)$, where $Y \in \mathbf{R}^{ M \times H' \times W'}$. We can further view $K$ as $M$ different filters, $\{K_i \in \mathbf{R}^{C \times k \times k}\}$. Since convolution is linear, we can rewrite $\text{Conv}(K,X)$ in a matrix-vector form:
\begin{align}
\label{eq:toeplitz}
Y = \text{Conv}(K, X) \Leftrightarrow  \mathbf{y} = \mathcal{K} \mathbf{x}
\end{align}
where $\mathbf{x}$ is $X$ flattened to a vector. Note that we adopt rigorous notations here while $\mathbf{x}$ and $X$ are not distinguished previously. Each row of $\mathcal{K}$ has non-zero entries corresponding to a particular filter $K_i$ at a particular spatial location. As a result, $\mathcal{K}$ can be constructed as a doubly block-Toeplitz (DBT) matrix $\mathcal{K} \in \mathbf{R}^{ (M H' W')\times (C H W)}$ from kernel tensor $K \in \mathbf{R}^{M \times C \times k \times k}$.

We can obtain the output tensor $Y$ by reshaping vector $\mathbf{y}$ back to the tensor form. Fig.\ref{fig:cconv} depicts an example of a convolution based on DBT matrix, where we have input size of $C\times4\times4$, kernel size of $M\times C\times 2 \times 2$ and stride 1. 


\subsection{Convolutional Orthogonality}
\label{sec:conv_orth}

Depending on the configuration of each layer, the corresponding matrix $\mathcal{K} \in \mathbf{R}^{ (M H' W')\times (C H W)}$ may be a fat matrix ($MH'W' \leq CHW$) or a tall matrix ($MH'W' > CHW$). In either case, we want to regularize the spectrum of $\mathcal{K}$ to be uniform. In the fat matrix case, the uniform spectrum requires a row orthogonal convolution, while the tall matrix case requires a column orthogonal convolution, where $\mathcal{K}$ is a normalized frame \cite{kovavcevic2008introduction} and preserves the norm.

In theory, we can implement the doubly block-Toeplitz matrix $\mathcal{K}$ and enforce the orthogonality condition in a brute force fashion. However, since $\mathcal{K}$ is highly structured and sparse, a much more efficient algorithm exists. In the following, we show the equivalent conditions to the row and column orthogonality, which can be easily computed.


\vspace{0.05in}
\noindent \textbf{Row Orthogonality.} As we mentioned earlier, each row of $\mathcal{K}$ corresponds to a filter $K_{i}$ at a particular spatial location $(h',w')$ flattened to a vector, denoted as $\mathcal{K}_{ih'w',\bigcdot} \in \mathbf{R}^{C H W}$. The row orthogonality condition is:
\begin{align}
\label{eq:roworthogonal}
\hspace{-0.10in} \langle \mathcal{K}_{ih_{1}'w_{1}',\bigcdot}~, \mathcal{K}_{jh_2'w_2',\bigcdot} \rangle = 
\begin{cases}
                                    1, (i,h_1',w_1') = (j,h_2',w_2')\\
                                    0, \text{otherwise}\\
                            \end{cases} \hspace{-0.10in}
\end{align}

In practice, we do not need to check pairs when the corresponding filter patches do not overlap. It is clear that $\langle \mathcal{K}_{ih_{1}'w_{1}',\bigcdot}~,\mathcal{K}_{jh_2'w_2',\bigcdot} \rangle = 0$ if either $|h_1 - h_2|\geq k$ or $|w_1 - w_2|\geq k$, since the two flattened vectors have no support overlap and thus have a zero inner product.  Thus, we only need to check Condition~\ref{eq:roworthogonal} where $|h_1 - h_2|, |w_1 - w_2| < k$.  Due to the spatial symmetry, we can choose fixed $h_1, w_1$ and only vary $i,j, h_2, w_2$, where $|h_1 - h_2|, |w_1 - w_2| < k$.

\begin{figure}[t!]
    \vspace{-0.1in}
        \includegraphics[width=0.47\textwidth]{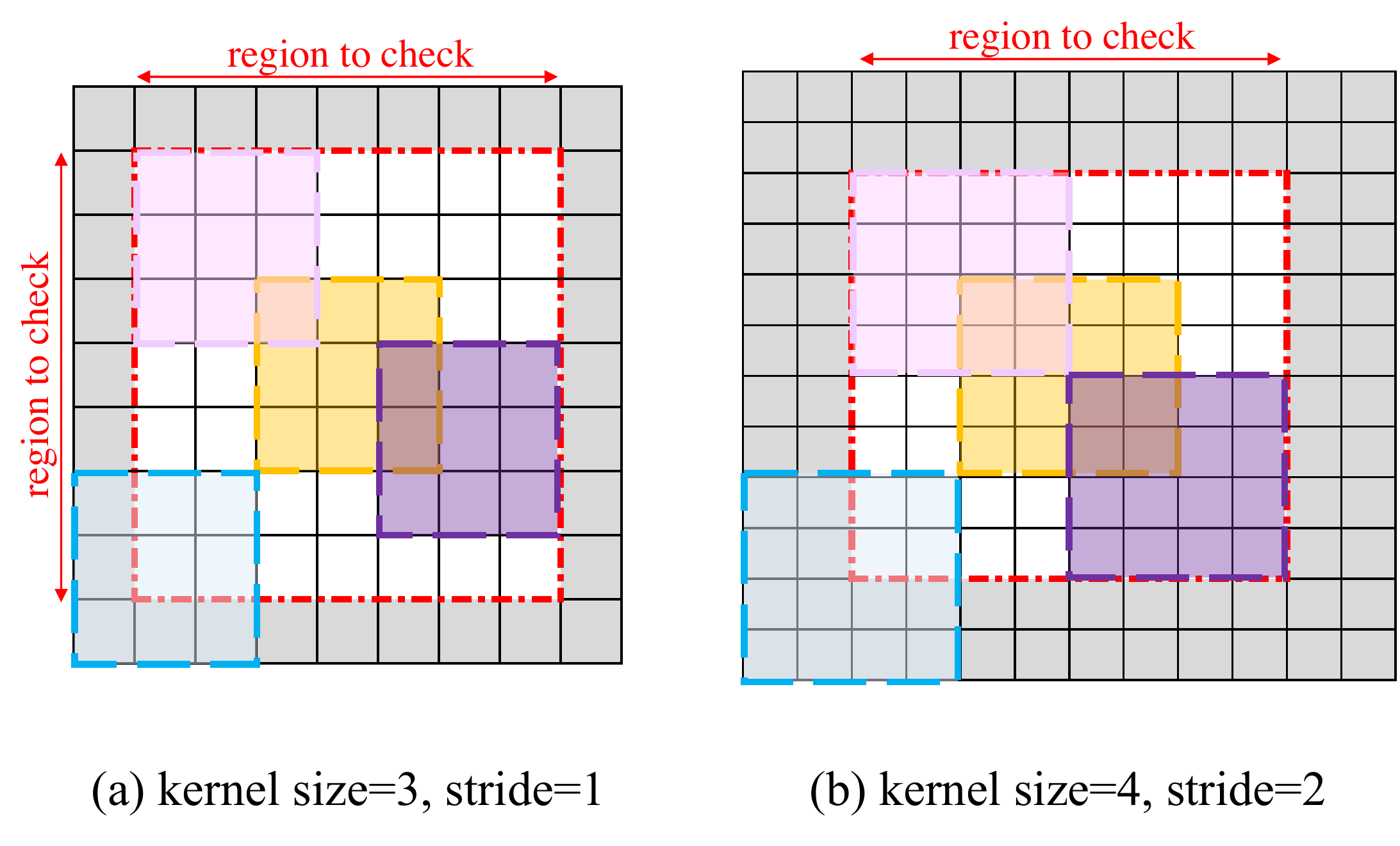}%
   \caption{\small The spatial region to check for row orthogonality. It is only necessary to check overlapping filter patches for the row orthogonality condition. We show two example cases: stride $S=1$ with kernel size $k=3$ and stride $S=2$ with kernel size $k=4$. In both examples, the orange patch is the center patch, and the red border is the region of overlapping patches. For example, pink and purple patches fall into the red region and overlap with the center region; blue patches are not fully inside the red region and they do not overlap with the orange ones. We can use padding to obtain the overlapping regions.}
   \label{fig:row}
   \vspace{-1em}
   \end{figure}

Fig.\ref{fig:row} shows examples of regions of overlapping filter patches. For a convolution with the kernel size $k$ and the stride $S$, the region to check orthogonality can be realized by the original convolution with padding $P = \lfloor \frac{k-1}{S} \rfloor \cdot S$.
Now we have an equivalent condition to Condition~\ref{eq:roworthogonal} as the following self-convolution:
\begin{align}
\label{eq:roworthogonal_conv2}
\text{Conv}(K, K, \text{padding} = P, \text{stride} = S) = I_{r0}
\end{align}
where $I_{r0} \in \mathbf{R}^{M\times M\times (2P/S+1) \times (2P/S +1)}$ is a tensor, which has zeros entries except for center $M\times M$ entries as an identity matrix. Minimizing the difference between $Z = \text{Conv}(K, K, \text{padding} = P, \text{stride} = S)$ and $I_{r0}$ gives us a near row-orthogonal  convolution in terms of DBT matrix $\mathcal{K}$.




\vspace{0.05in}
\noindent \textbf{Column Orthogonality.} We use tensor $E_{i,h,w} \in \mathbf{R}^{C \times H \times W}$ to denote an input tensor, which has all zeros except a 1 entry at the $i^{\textbf{th}}$ input channel, spatial location $(h,w)$. Denoting $\mathbf{e_{ihw}}\in \mathbf{R}^{C H W}$ as the flattened vector of $E_{i,h,w}$, we can obtain a column $\mathcal{K}_{\kern 0.1em \bigcdot,ihw}$ of $\mathcal{K}$ by multiply $\mathcal{K}$ and vector $\mathbf{e_{ihw}}$:
\begin{align}
\label{eq:row}
    \mathcal{K}_{\kern 0.1em\bigcdot, ihw} = \mathcal{K} \mathbf{e_{ihw}} = \text{Conv}(K, E_{i,h,w})
\end{align}
Here, we slightly abuse the equality notation as the reshaping is easily understood. The column orthogonality condition is:
\begin{align}
\label{eq:colorthogonal}
\hspace{-0.1in}\langle \mathcal{K}_{\kern 0.1em \bigcdot, ih_1w_1}~, \mathcal{K}_{\kern 0.1em \bigcdot, jh_2w_2} \rangle = 
\begin{cases}
                                    1, (i,h_1,w_1) = (j,h_2,w_2)\\
                                    0,\text{otherwise}\\
                            \end{cases}\hspace{-0.1in}
\end{align}

Similar to the row orthogonality, since the spatial size of $K$ is only $k$, Condition \ref{eq:colorthogonal} only needs to be checked in a local region where there is spatial overlap between $\mathcal{K}_{\kern 0.1em \bigcdot, ih_1w_1}$ and $\mathcal{K}_{\kern 0.1em \bigcdot, jh_2w_2}$. For the stride 1 convolution case, there exists a simpler condition equivalent to Condition~\ref{eq:colorthogonal}:
\begin{align}
\label{eq:colorthogonal_conv}
\text{Conv}(K^T, K^T, \text{padding} = k-1, \text{stride} = 1) = I_{c0}
\end{align}
where $K^T$ is the input-output transposed $K$, i.e. $K^T \in \mathbf{R}^{C\times M\times k\times k}$. $I_{c0} \in \mathbf{R}^{C\times C\times (2k-1) \times (2k-1)}$ has all zeros except for the center $C\times C$ entries as an identity matrix.


\vspace{0.05in}
\noindent \textbf{Comparison to Kernel Orthogonality.} The kernel row- and column-orthogonality conditions can be written in the following convolution form respectively:
\begin{align}
\label{eq:orthogonal_convkernel}
\begin{cases}
\text{Conv}(K, K, \text{padding} = 0) = I_{r0} \\
\text{Conv}(K^T, K^T, \text{padding} = 0) = I_{c0}
\end{cases}
\end{align}
where tensor $I_{r0} \in \mathbf{R}^{M\times M\times 1 \times 1}$, $I_{c0} \in \mathbf{R}^{C\times C\times 1 \times 1}$ are both equivalent to identity matrices\footnote{Since there is only 1 spatial location.}.

Obviously, the kernel orthogonality conditions~\ref{eq:orthogonal_convkernel} are necessary but not sufficient conditions for the orthogonal convolution conditions~\ref{eq:roworthogonal_conv2},\ref{eq:colorthogonal_conv} in general. For the special case when convolution stride is $k$, they are equivalent.

\vspace{0.05in}
\noindent \textbf{Row-Column Orthogonality Equivalence.} The lemma below unifies the row orthogonality condition~\ref{eq:roworthogonal} and column orthogonality condition~\ref{eq:colorthogonal}. This lemma \cite{le2011ica} gives a uniform convolution orthogonality independent of the actual shape of $\mathcal{K}$ and provides a unique regularization: $\min_{K} L_{\text{orth}} = \|Z-I_{r0} \|_F^2$, which only depends on Condition~\ref{eq:roworthogonal_conv2}.

\begin{lemma}
\label{lemma:equivalence}
The row orthogonality and column orthogonality are equivalent in the MSE sense, i.e. $\|\mathcal{K}\mathcal{K}^T - I\|_F^2 = \|\mathcal{K}^T\mathcal{K} - I^\prime\|_F^2 + U$, where $U$ is a constant.
\end{lemma}
\noindent We leave the proof to Section~\ref{sec:proof} of supplementary materials.




\vspace{0.05in}
\noindent \textbf{Orthogonal Regularization in CNNs.} We add an additional soft orthogonal convolution regularization loss to the final loss of CNNs, so that the task objective and orthogonality regularization can be simultaneously achieved. Denoting $\lambda>0$ as the weight of the orthogonal regularization loss, the final loss is:
\begin{align}
    L = L_{\text{task}} + \lambda L_{\text{orth}}
\end{align}
where $L_{\text{task}}$ is the task loss, e.g. softmax loss for image classification, and $L_{\text{orth}}$ is the orthogonal regularization loss.

%% file: sections/4experiments.tex
\section{Experiments}
\label{sec:exp}
We conduct 3 sets of experiments to evaluate OCNNs.  The first set benchmarks our approach on image classification datasets CIFAR100 and ImageNet.  The second set benchmarks the performance under semi-supervised settings and focuses on qualities of learned features.  For high-level visual feature qualities, we experiment on the fine-grained bird image retrieval.  For low-level visual features, we experiment on unsupervised image inpainting. Additionally, we compare visual feature qualities in image generation tasks.  The third set of experiments focuses on the robustness of OCNN under adversarial attacks.  We analyze OCNNs in terms of DBT matrix $\mathcal{K}$'s spectrum, feature similarity, hyperparameter tuning, and space/time complexity.

\subsection{ Classification on CIFAR100}
The key novelty of our approach is the orthogonal regularization term on convolutional layers.  We compare both conv-orthogonal and kernel-orthogonal regularizers on CIFAR-100 \cite{krizhevsky2009learning} and evaluate the image classification performance using ResNet \cite{he2016deep} and WideResNet \cite{zagoruyko2016wide} as backbone networks. The kernel-orthogonality and our conv-orthogonality are added as additional regularization terms, without modifying the network architecture. Hence, the number of parameters of the network does not change.

\vspace{0.05in}
\noindent\textbf{ResNet and Row Orthogonality.} Though we have derived a unified orthogonal convolution regularizer, we benchmark its effectiveness with two different settings. Convolutional layers in ResNet \cite{he2016deep} usually preserve or reduce the dimension from input to output, \ie a DBT matrix $\mathcal{K}$ would be a square or fat matrix. In this case, our regularizer leads to the row orthogonality condition. Table \ref{tab:cifar} shows top-1 classification accuracies on CIFAR100. Our approach achieves 78.1\%, 78.7\%, and 79.5\% image classification accuracies with ResNet18, ResNet34 and ResNet50, respectively. For 3 backbone models, OCNNs outperform plain baselines by 3\%, 2\%, and 1\%, as well as kernel orthogonal regularizers by 2\%, 1\%, and 1\%. 

\vspace{0.05in}
\noindent\textbf{WideResNet and Column Orthogonality.} Unlike ResNet, WideResNet \cite{zagoruyko2016wide} has more channels and some tall DBT matrices $\mathcal{K}$. When the corresponding DBT matrix $\mathcal{K}$ of a convolutional layer increases dimensionality from the input to the output, our OCNN leads to the column orthogonality condition. Table \ref{tab:wr} reports the performance of column orthogonal regularizers with backbone model of WideResNet28 on CIFAR100. Our OCNNs achieve 3\% and 1\% gain over plain baselines and kernel orthogonal regularizers.

   

\begin{table}[!ht]
\begin{center}
\centering
\caption{\small Top-1 accuracies on CIFAR100. Our OCNN outperforms baselines and the SOTA orthogonal regularizations.}
\label{tab:cifar}
\resizebox{\columnwidth}{!}{
\begin{tabular}{>{\columncolor[gray]{0.95}}c|c|c|c}
\hline
\rowcolor{LightCyan} & ResNet18 & ResNet34 & ResNet50 \\ \hline
baseline \cite{he2016deep}                  & 75.3     & 76.7     & 78.5     \\ \hline
kernel orthogonality \cite{xie2017all}      & 76.5     & 77.5     & 78.8     \\ \hline
OCNN (ours) & \textbf{78.1}     & \textbf{78.7}     & \textbf{79.5}     \\ \hline
\end{tabular}

}
 \vspace{-0.3in}

\end{center}
\end{table}

\begin{table}[!ht]

\begin{center}
\centering
\caption{\small WideResNet \cite{zagoruyko2016wide} performance. We observe improved performance of OCNNs.}
\label{tab:wr}

\resizebox{0.8\columnwidth}{!}{
\begin{tabular}{>{\columncolor[gray]{0.95}}c|c|c|c}
\hline
\rowcolor{LightCyan}& WideResNet \cite{zagoruyko2016wide}& Kernel orth \cite{xie2017all}& OCNN \\ \hline
Acc. & 77.0       & 79.3        & \textbf{80.1} \\ \hline
\end{tabular}
}
 \vspace{-0.34in}
\end{center}
\end{table}

\subsection{ Classification on ImageNet}
\label{sec:imagenet}
We add conv-orthogonal regularizers to the backbone model ResNet34 on ImageNet \cite{deng2009imagenet}, and compare OCNNs with state-of-the-art orthogonal regularization methods. 

\vspace{0.05in}
\noindent \textbf{Experimental Settings.} We follow the standard training and evaluation protocols of ResNet34. In particular, the total epoch of the training is 90. We start the learning rate at 0.1, decreasing by 0.1 every 30 epochs and weight decay 1e-4. The weight $\lambda$ of the regularization loss is 0.01, the model is trained using SGD with momentum 0.9, and the batch size is 256.

\vspace{0.05in}
\noindent \textbf{Comparisons.} Our method is compared with hard orthogonality OMDSM \cite{huang2018orthogonal}, kernel orthogonality \cite{xie2017all} and spectral restricted isometry property regularization \cite{bansal2018can}. Table \ref{tab:imagenet} shows the Top-1 and Top-5 accuracies on ImageNet. Without additional modification to the backbone model, OCNN achieves 25.87\% top-5 and 7.89\% top-1 error. The proposed method outperforms the plain baseline, as well as other orthogonal regularizations by 1\%.

\begin{table}[!ht]
\begin{center}
\centering
\caption{\small Top-1 and Top-5 errors on ImageNet \cite{deng2009imagenet} with ResNet34 \cite{he2016deep}. Our conv-orthogonal regularization outperforms baselines and SOTA orthogonal regularizations.}
\label{tab:imagenet}

\resizebox{0.85\columnwidth}{!}{
\begin{tabular}{>{\columncolor[gray]{0.95}}c|c|c}
\hline
\rowcolor{LightCyan}                   & Top-1 error & Top-5 error \\
ResNet34 (baseline) \cite{he2016deep}         & 26.70       & 8.58        \\ \hline
OMDSM \cite{huang2018orthogonal} & 26.88           & 8.89        \\ \hline
kernel orthogonality \cite{xie2017all} & 26.68           & 8.43        \\ \hline
SRIP \cite{bansal2018can}             & 26.10           & 8.32        \\ \hline
OCNN (ours)              &     \textbf{25.87}     & \textbf{7.89}  \\ \hline
\end{tabular}
}
\vspace{-0.25in}

\end{center}
\end{table}

\subsection{Semi-Supervised Learning}

A general regularizer should provide benefit to a variety of tasks. A common scenario that benefits from regularization is semi-supervised learning, where we have a large amount of data with limited labels. We randomly sample a subset of CIFAR100 as labeled and treat the rest as unlabeled. The orthogonal regularization is added to the baseline model ResNet18 without any additional modifications. The classification performance is evaluated on the entire validation set for all different labeled subsets.

We compare OCNN with kernel-orthogonal regularization while varying the proportion of labeled data from 10\% to 80\% of the entire dataset (Table \ref{tab:semi}). OCNN constantly outperforms the baseline by 2\% - 3\% under different fractions of labeled data.
 
\begin{table}[!ht]
\begin{center}
\centering
\caption{\small Top-1 accuracies on CIFAR100 with different fractions of labeled data. OCNNs are consistently better.}
\label{tab:semi}

\resizebox{\columnwidth}{!}{
\begin{tabular}{>{\columncolor[gray]{0.95}}c|c|c|c|c|c|c}
\hline
\rowcolor{LightCyan}        \% of training data          & 10\% & 20\% & 40\% & 60\% & 80\% & 100\% \\ \hline
ResNet18    \cite{he2016deep}        & 31.2 & 47.9 & 60.9 & 66.6 & 69.1 & 75.3  \\ \hline
kernel orthogonality \cite{xie2017all} & 33.7 & 50.5 & 63.0 & 68.8 & 70.9 & 76.5  \\ \hline
Conv-orthogonality   & \textbf{34.5} & \textbf{51.0} & \textbf{63.5} & \textbf{69.2} & \textbf{71.5} & \textbf{78.1}  \\ \hline
Our gain           & 3.3  & 3.1  & 2.6 & 2.6  & 2.4  & 2.8   \\ \hline
\end{tabular}
}
\vspace{-0.4in}

\end{center}
\end{table}

\subsection{Fine-grained Image Retrieval}
\begin{figure}[t!]
        \includegraphics[width=\columnwidth]{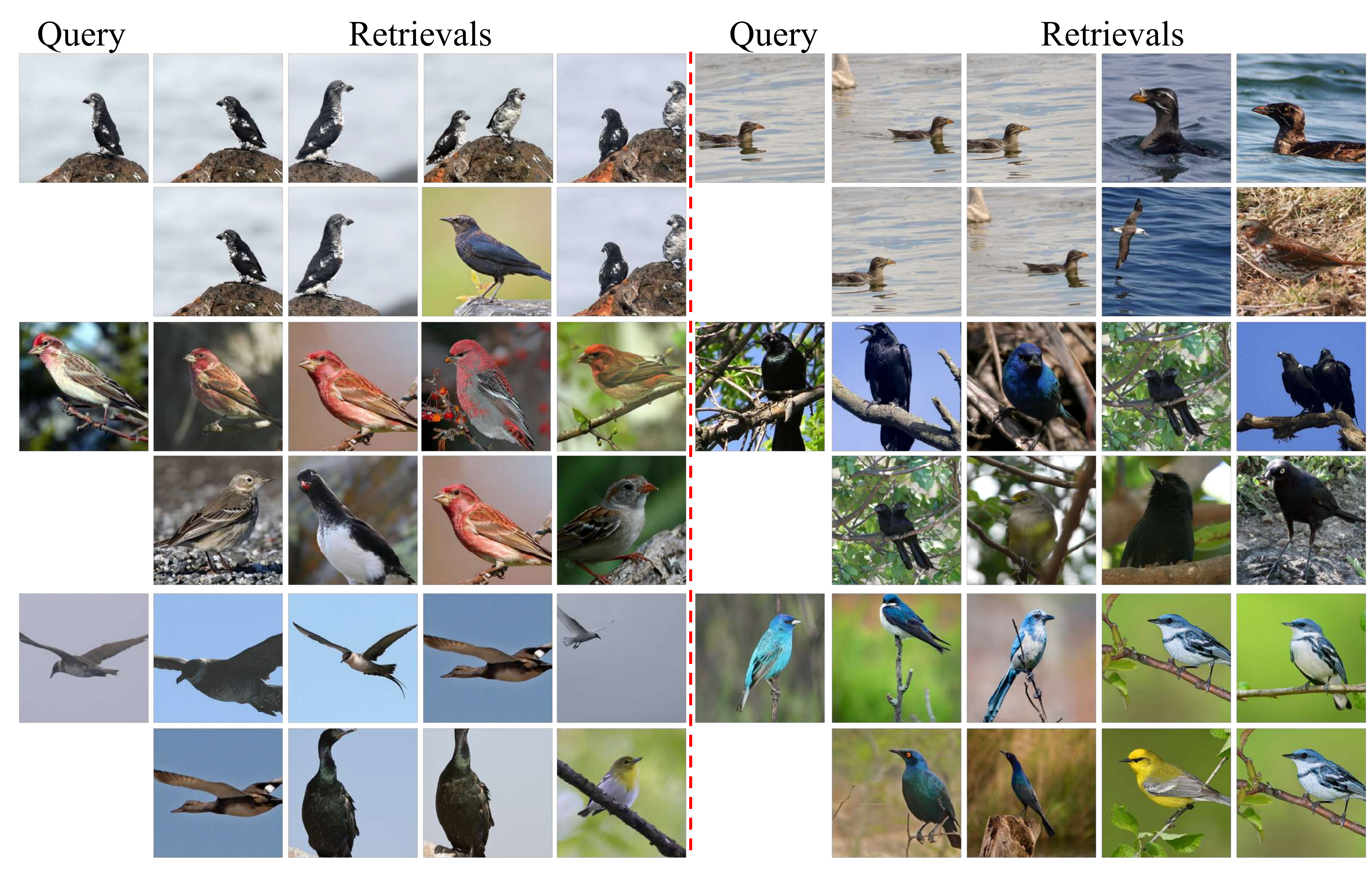}%
   \caption{\small Image retrieval results on CUB-200 Birds Dataset. The model (ResNet34) is trained on ImageNet only. First row shows our OCNN results, while the second row shows the baseline model results. Ours achieves 2\% and 3\% top-1 and top-5 $k$-nearest neighbor classification gain.}
   \label{fig:ret}
   \vspace{-1em}
   \end{figure}
We conduct fine-grained image retrieval experiments on CUB-200 bird dataset \cite{WelinderEtal2010} to understand  high-level visual feature qualities of OCNNs. Specifically, we directly use the ResNet34 model trained on ImageNet (from Section \ref{sec:imagenet}) to obtain features of images in CUB-200, without further training on the dataset. We observed improved results with OCNNs (Fig.\ref{fig:ret}). With conv-orthogonal regularizers, the top-1 $k$-nearest-neighbor classification accuracy improves from 25.1\% to 27.0\%, and  top-5 $k$-nearest-neighbor classification accuracy improves from 39.4\% to 42.3\%.

\subsection{Unsupervised Image Inpainting}

\begin{figure}[t]
        \includegraphics[width=\columnwidth]{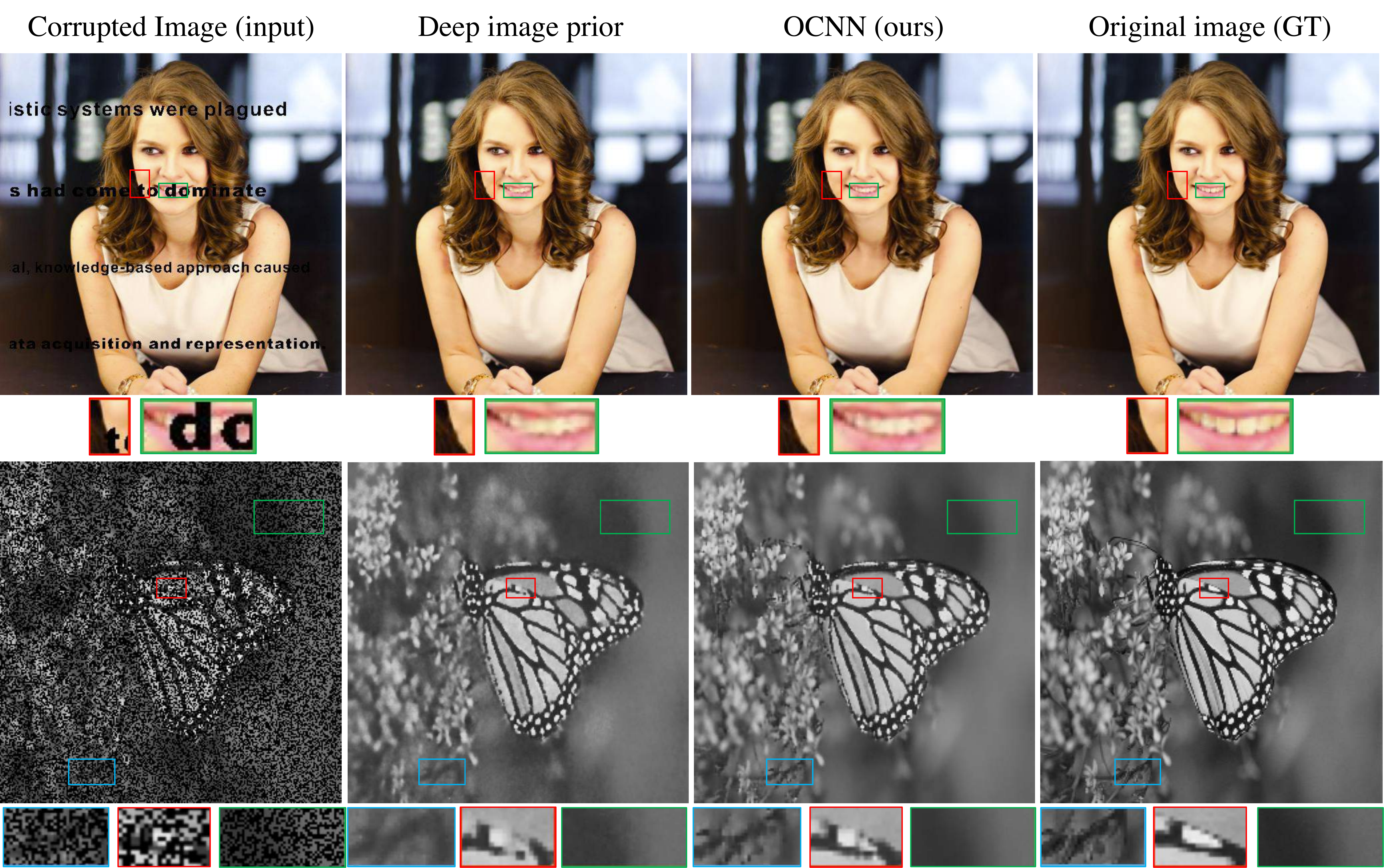}%
   \caption{\small Image inpainting results compared with deep image prior \cite{Ulyanov_2018_CVPR}. Top – comparison on text inpainting example. Bottom – comparison on inpainting 50\% of missing pixels. In both cases, our approach outperforms previous methods.}
   \label{fig:inp}
\vspace{-0.12in}
\end{figure}

\begin{table*}[!ht]
\begin{center}
\centering
\caption{\small  Quantitative comparisons on the standard inpainting dataset \cite{Heide_2015_CVPR}. Our conv-orthogonality outperforms the SOTA methods.}
\vspace{-0.05in}
\label{tab:inp}
\resizebox{1.0\textwidth}{!}{
\begin{tabular}{>{\columncolor[gray]{0.95}}c|c|c|c|c|c|c|c|c|c|c|c}
\hline
\rowcolor{LightCyan}    & Barbara & Boat & House & Lena & Peppers & C.man & Couple & Finger & Hill & Man & Montage \\ \hline
Convolutional dictionary learning \cite{Papyan_2017_ICCV} &   28.14 &31.44& 34.58& 35.04 &31.11& 27.90& 31.18& 31.34 &32.35& 31.92 & 28.05         \\ \hline
Deep image prior (DIP) \cite{Ulyanov_2018_CVPR}           &       32.22 &33.06 &39.16 &36.16 &33.05 &29.8 &32.52 &32.84 &32.77& 32.20 &34.54        \\ \hline
DIP + kernel orthogonality \cite{xie2017all}       &    34.88     &  34.93    &   38.53    &   37.66   &  34.58       &   33.18    &   33.71     &  34.40   &  35.98  &  32.93  &   36.99   \\ \hline
DIP + conv-orthogonality (ours)        &  \textbf{38.12}&\textbf{35.15}&\textbf{41.73}&\textbf{39.76}&\textbf{37.75}&\textbf{38.21}&\textbf{35.88}&\textbf{36.87}&\textbf{39.89}&\textbf{33.57}&\textbf{38.48}
     \\ \hline
\end{tabular}
}
\vspace{-0.25in}

\end{center}
\end{table*}

To further assess the generalization capacity of OCNNs, we add the regularization term to the new task of unsupervised inpainting. In image inpainting, one is given an image $X_0$ with missing pixels in correspondence of a binary mask $M \in \{ 0, 1\}^{C\times H \times W}$ of the same size of the image. The goal is to reconstruct the original image $X$ by recovering missing pixels:
\begin{align}
\label{eq:inp}
\min E(X; X_0) = \min \|(X-X_0) \odot M\|^2_F
\end{align}

Deep image prior (DIP) \cite{Ulyanov_2018_CVPR} proposed to use the prior implicitly captured by the choice of a particular generator network  $f_{\theta}$ with parameter $\theta$. Specifically, given a code vector/ tensor $\textbf{z}$, DIP used CNNs as a parameterization $X = f_{\theta}(\mathbf{z})$. The reconstruction goal in Eqn.\ref{eq:inp} can be written as:
\begin{align}
 \min_{\theta} \|(f_{\theta}(\mathbf{z})-X_0) \odot M\|^2_F
\end{align}

The network can be optimized without training data to recover $X$. We further add our conv-orthogonal regularization as an additional prior to the reconstruction goal, to validate if the proposed regularization helps the inpainting:
\begin{align}
 \min_{\theta} \|(f_{\theta}(\mathbf{z})-X_0) \odot M\|^2_F + \lambda L_{\text{orth}}(\theta)
\end{align}
\vspace{-0.15in}

In the first example (Fig.\ref{fig:inp}, top), the inpainting is used to remove text overlaid on an image. Compared with DIP \cite{Ulyanov_2018_CVPR}, our orthogonal regularization leads to improved reconstruction result of details, especially for the smoothed face outline and finer teeth reconstruction.

The second example (Fig.\ref{fig:inp}, bottom) considers inpainting with masks randomly sampled according to a binary Bernoulli distribution. Following the procedure in \cite{Papyan_2017_ICCV,Ulyanov_2018_CVPR}, we sample a mask to randomly drop 50\% of pixels. For a fair comparison, all the methods adopt the same mask. We observe improved background quality, as well as finer reconstruction of the texture of butterfly wings.

We report quantitative PSNR comparisons on the standard data set \cite{heide2015fast} in Table \ref{tab:inp}. OCNN outperforms previous state-of-the-art DIP \cite{Ulyanov_2018_CVPR} and convolutional sparse coding \cite{Papyan_2017_ICCV}. We also observe performance gains compared to kernel orthogonal regularizations.


\subsection{Image Generation}

\label{sec:gen}

\begin{figure}[!t]
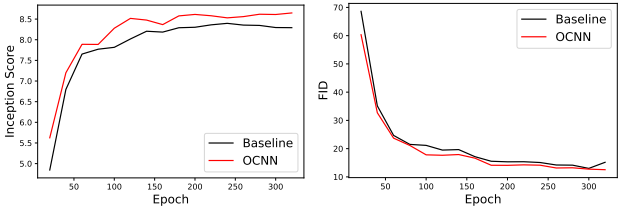

        \includegraphics[width=0.5\columnwidth]{figures/GAN_IS.pdf}%
        \includegraphics[width=0.5\columnwidth]{figures/GAN_FID.pdf}%
       \caption{\small  OCNNs have faster convergence for GANs. For IS (left) and FID (right), OCNNs consistently outperforms CNNs \cite{gong2019autogan} at every epoch.}
\label{fig:fastgan}
 \vspace{-0.2in}
\end{figure}

Orthogonal regularizers have shown great success in improving the stability and performance of GANs \cite{brock2016neural, miyato2018spectral, brock2018large}. We analyze the influence of convolutional orthogonal regularizers on GANs with the best architecture reported in \cite{gong2019autogan}. Training takes 320 epochs with OCNN regularizer applied to both the generator and discriminator. The regularizer loss $\lambda$ is set to 0.01 while other settings are retained as  default. 

\begin{table}[h]
\begin{center}
\centering
\caption{\small Inception Score and Fréchet Inception Distance comparison on CIFAR10. Our OCNN outperforms the baseline \cite{gong2019autogan} by 0.3 IS and 1.3 FID.}
\vspace{0.1in}
\label{tab:gen}

\resizebox{0.24\textwidth}{!}{
\begin{tabular}{>{\columncolor[gray]{0.95}}c|c|c}
\hline
\rowcolor{LightCyan}       & IS   & FID   \\
PixelCNN \cite{van2016conditional} & 4.60 & 65.93 \\
PixelIQN \cite{pmlr-v80-ostrovski18a} & 5.29 & 49.46 \\
EBM \cite{du2019implicit}     & 6.78 & 38.20 \\
SNGAN \cite{miyato2018spectral}   & 8.22 & 21.70 \\
BigGAN \cite{brock2018large}  & \textbf{9.22} & 14.73 \\
AutoGAN \cite{gong2019autogan} & 8.32 & 13.01 \\ \hline
OCNN (ours)     & 8.63 & \textbf{11.75}  \\ \hline
\end{tabular}
}
 \vspace{-1.5em}
\end{center}
\end{table}

The reported model is evaluated 5 times with 50k images each.
We achieve an inception score (IS) of $8.63 \pm 0.007$  and Fréchet inception distance (FID) of $11.75 \pm 0.04$   (Table \ref{tab:gen}),  outperforming the baseline and achieving the state-of-the-art performance. Additionally, we observe faster convergence of GANs with our regularizer (Fig.\ref{fig:fastgan}).
   
\subsection{Robustness under Attack}
\label{sec:attack}

The uniform spectrum of $\mathcal{K}$ makes each convolutional layer approximately a $1$-Lipschitz function. Given a small perturbation to the input, $\Delta x$, the change of the output $\Delta y$ is bounded to be low. Therefore, the model enjoys robustness under attack. Our experiments demonstrate that it is much harder to search for adversarial examples. 
   
\begin{table}[!t]
\begin{center}
\centering
\caption{\small Attack time and number of necessary attack queries needed for 90\% successful attack rate.}
\label{tab:attack}

\resizebox{\columnwidth}{!}{
\begin{tabular}{>{\columncolor[gray]{0.95}}c|c|c}
\hline
\rowcolor{LightCyan}       &  Attack time/s & \# necessary attack queries  \\ \hline
ResNet18 \cite{he2016deep}   &      19.3        &  27k          \\ \hline
OCNN (ours) &    136.7         &    46k        \\ \hline
\end{tabular}
}
 \vspace{-1.5em}
\end{center}
\end{table}

\begin{figure}[t!]
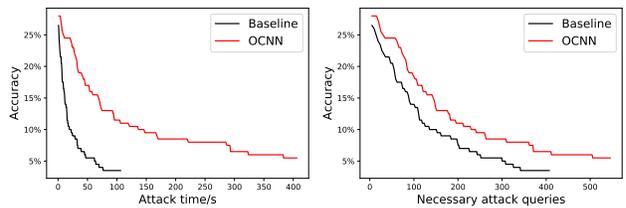

        \includegraphics[width=0.5\columnwidth]{figures/attack_time.pdf}%
        \includegraphics[width=0.5\columnwidth]{figures/attack.pdf}%
       \caption{\small  Model accuracy v.s. attack time and necessary attack queries. With our conv-orthogonal regularizer, it takes  7x time and 1.7x necessary attack queries to achieve 90\% successful attack rate. Note that baseline ends at accuracy 3.5\% while ours ends at 5.5\% with the same iteration.}
\label{fig:attack}
 \vspace{-1em}

   \end{figure}
We adopt the simple black box attack \cite{guo2019simple} to evaluate the robustness of baseline and OCNN with ResNet18 \cite{he2016deep} backbone architecture trained on CIFAR100. The attack samples around the input image and finds a ``direction'' to rapidly decrease the classification confidence of the network by manipulating the input. We only evaluate on the correctly classified test images. The maximum iteration is 10,000 with pixel attack. All other settings are retained. We report the attack time and number of necessary attack queries for a specific attack successful rate.

It takes approximately 7x time and 1.7x attack queries to attack OCNN, compared with the baseline (Table \ref{tab:attack} and Fig.\ref{fig:attack}  ). Additionally, after the same iterations of the attack, our model outperforms the baseline by 2\%.

To achieve the same attack rate, baseline models need more necessary attack queries, and searching for such queries is nontrivial and time consuming. This may account for the longer attack time of the OCNN.

\subsection{Analysis}
\label{sec:exp_analysis}
To understand how the conv-orthogonal regularization help improve the performance of CNNs, we analyze several aspects of OCNNs. First, we analyze the spectrum of the DBT matrix $\mathcal{K}$ to understand how it helps relieve gradient vanishing/exploding. We then analyze the filter similarity of networks over different layers, followed by the influence of the weight $\lambda$ of the regularization term. Finally, we analyze the time and space complexity of OCNNs.

\vspace{0.05in}
\noindent \textbf{Spectrum of the DBT Kernel Matrix $\mathcal{K}$.} For a convolution $Y = \mathcal{K}X$, we analyze the spectrum of $\mathcal{K}$ to understand the properties of the convolution. We analyze the spectrum of $K \in \mathbf{R}^{64\times128\times3\times3}$ of the first convolutional layer of the third convolutional block of ResNet18 network trained on CIFAR100. For fast computation, we use input of size $64 \times 16\times 16 $, and solve all the singular values of $\mathcal{K}$.

As in Fig.\ref{fig:intro}(b), the spectrum of plain models vanishes rapidly, and may cause gradient vanishing problems. Kernel orthogonality helps the spectrum decrease slower. With our conv-orthogonal regularization, the spectrum almost always stays at 1. The uniform spectrum perserves norm and information between convolutional layers.

\vspace{0.05in}
 \noindent \textbf{Filter Similarity}. Orthogonality  makes  off-diagonal elements become 0. This means that for any two channels of a convolutional layer, correlations should be relatively small. This can reduce the filter similarity and feature redundancy across different channels.
 
 We use guided back-propagation patterns \cite{springenberg2015striving} on images from the validation set of the ImageNet \cite{deng2009imagenet} dataset to investigate filter similarities. Guided back-propagation patterns visualize the gradient of a particular neuron with respect to the input image $X \in \mathbf{R}^{C \times H \times W}$. Specifically for a layer of $M$ channels, the combination of flattened guided back-propagation patterns is denoted as $G\in \mathbf{R}^{M \times C W H}$. The correlation matrix $\text{corr}(G)$ over different channels of this layer is $ (\text{diag} (K_{GG}))^{-\frac{1}{2}} K_{GG}(\text{diag} (K_{GG}))^{-\frac{1}{2}}$,
where $K_{GG} = \frac{1}{M}[(G-\E[G])(G-\E[G])^T]$ is the covariance matrix. We plot the histogram of off-diagonal elements of $\text{corr}(G)$ of all validation images.

Fig.\ref{fig:intro}(a) depicts the normalized histogram of pairwise filter similarities of plain ResNet34.  As the number of channels increases with depth from 128 to 512, the curve shifts right and becomes far narrower, i.e., more filters become similar. Fig.\ref{fig:intro}(c) depicts the histogram of filter similarities at layer 27 of ResNet34 with different regularizers. OCNNs make the curve shift left and become wider, indicating it can enhance filter diversity and decrease feature redundancy.
 
\begin{figure}
\centering
       \includegraphics[width=0.5\textwidth]{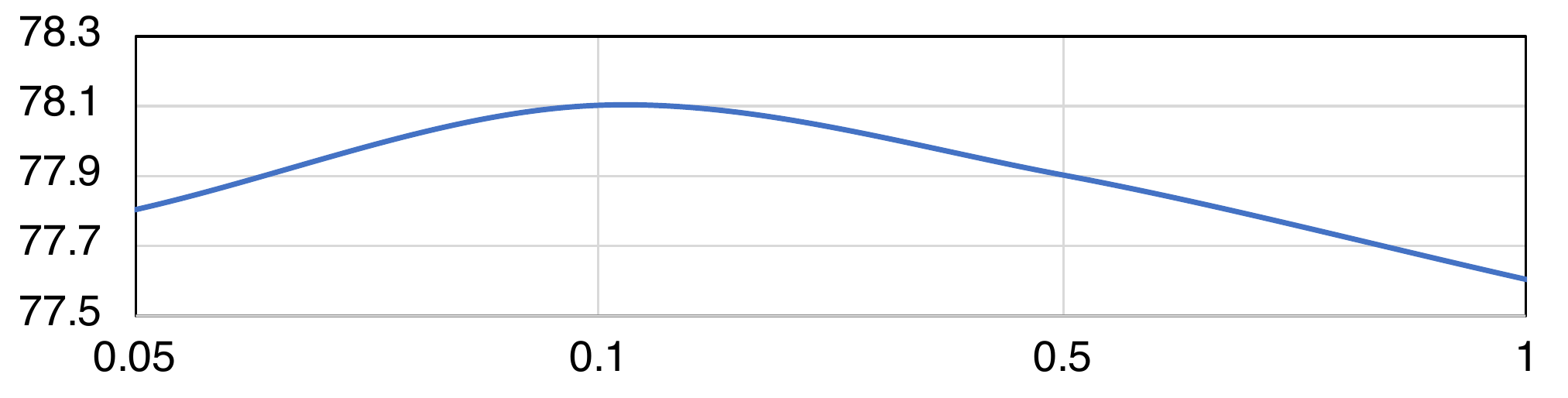}%
   \caption{\small CIFAR100 classification accuracy (\%) with different weight $\lambda$ of the regularization loss.  With backbone model ResNet18, we achieve the highest  performance at $\lambda=0.1$.}
   \vspace{-1em}
   \label{fig:hyper}
   \end{figure}
   
\vspace{0.05in}
\noindent \textbf{Hyper-Parameter Analysis.} We analyze the influence of the weight $\lambda$ of the  orthogonality loss. As discussed earlier, we achieve the ``soft orthogonality'' by adding additional loss with weight $\lambda$ to the network. Fig.\ref{fig:hyper} plots the image classification performance of CIFAR100 with backbone model ResNet18 under $\lambda$ ranging from 0.05 to 1.0. Our approach achieves the highest   accuracy when $\lambda=0.1$.

\vspace{0.05in}
\noindent \textbf{Space and Time Complexity}. We analyze the space and time complexity in Table \ref{tab:complexity}. The ResNet34 \cite{he2016deep} backbone model is tested on ImageNet \cite{deng2009imagenet} with a single NVIDIA GeForce GTX 1080 Ti GPU and batch size 256.  

The number of parameters and the test time of the CNN do not change since the  regularizer is an additional loss term only used during training. With kernel orthogonal regularizers, the training time increases 3\%; with conv-orthogonal regularizers, the training time increases 9\%.

\begin{table}[!ht]
\caption{ Model size and training/ test time on ImageNet \cite{deng2009imagenet}.}
\vspace{-0.25in}
\begin{center}
\resizebox{\columnwidth}{!}{
\begin{tabular}{>{\columncolor[gray]{0.95}}c|c|c|c}
\hline
\rowcolor{LightCyan}                     & ResNet34 \cite{he2016deep}  &  kernel-orth \cite{xie2017all} & OCNN \\ \hline
\# Params.      & 21.8M   & same     & same                  \\ \hline
Training time (min/epoch) & 49.5   &  51.0    & 54.1             \\ \hline
Test time (min/epoch) &  1.5   &   same   &       same            \\ \hline
\end{tabular}
}
\vspace{-0.33in}

\end{center}

\label{tab:complexity}
\end{table}

%% file: sections/5conclusion.tex
\section{Summary}
We develop an efficient OCNN approach to impose a filter orthogonality condition on a convolutional layer based on the doubly block-Toeplitz matrix representation of the convolutional kernel, as opposed to the commonly adopted kernel orthogonality approaches.  We show that  kernel orthogonality \cite{bansal2018can, huang2018orthogonal} is  necessary but not sufficient for ensuring orthogonal convolutions.

Our OCNN requires no additional parameters and little computational overhead, consistently outperforming the state-of-the-art alternatives on a wide range of tasks such as  image classification and inpainting under supervised, semi-supervised and unsupervised settings.  It learns more diverse and expressive  features with better training stability, robustness, and generalization. 

{ \small \noindent{\bf Acknowledgements.} This research was supported, in part, by Berkeley Deep Drive, DARPA, and NSF-IIS-1718991.
The authors thank Xudong Wang for discussions on filter similarity, Jesse Livezey for the pointer to a previous proof for row-column orthogonality equivalence, Haoran Guo, Ryan Zarcone, and Pratik Sachdeva for proofreading, and anonymous reviewers for their insightful comments.}

%% file: append.tex
\section*{Supplementary Materials}
The supplementary materials provide intuitive explanations of our approach (Section \ref{sec:int}), network dissection results to understand the change in feature redundancy/expressiveness (Section \ref{sec:disec}), deep metric learning performance to understand the generalizability (Section \ref{sec:DML}), proof of Lemma \ref{lemma:equivalence} (Section \ref{sec:proof}) and visualizations of filter similarities (Section \ref{sec:feat_sim}).  


\renewcommand{\thesubsection}{\Alph{subsection}}

\subsection{Intuitive Explanations of our Approach}
\label{sec:int}

We analyze a convolutional layer which transforms input $X$ to output $Y$ with a learnable kernel $K$: $Y = \text{Conv}(K, X)$ in CNNs. Writing in linear matrix-vector multiplication form $Y = \mathcal{K} X$ (Fig.\ref{fig:overview}b of the paper), we simplify the analysis from the perspective of linear systems. We do not use \textit{im2col} form $Y =K \widetilde{X}$ (Fig.\ref{fig:overview}a of the paper) as there is an additional structured linear transform from $X$ to $\tilde{X}$ and this additional linear transform makes the analysis indirect. As we mentioned earlier, the kernel orthogonality does not lead to a uniform spectrum.

The spectrum of $\mathcal{K}$ reflects the scaling property of the corresponding convolutional layer: different input $X$ (such as cat, dog, and house images) would scale up by $\eta = \frac{\|Y\|}{\|X\|}$. The scaling factor $\eta$ also reflects the scaling of the gradient. A typical CNN has highly non-uniform convolution spectrum (Fig.\ref{fig:intro}b of the paper): for some inputs, it scales up to 2; for others, it scales by $0.1$. For a deep network, these irregular spectra are multiplied together and can potentially lead to gradient exploding and vanishing issues.

Features learned by CNNs are also more redundant due to the non-uniform spectrum issues (Fig.\ref{fig:intro}a of the paper). This comes from the diverse learning ability for different images and a uniform spectrum distribution could alleviate the problem. In order to achieve a uniform spectrum, We make convolutions orthogonal by enforcing DBT kernel matrix $\mathcal{K}$ to be orthogonal. The orthogonal convolution regularizer leads to uniform $\mathcal{K}$ spectra as expected. It further reduces  feature redundancy and improves the performance (Fig.\ref{fig:intro}b,c,d of the paper).

Besides classification performance improvements, we also observe improved visual feature qualities, both in high-level (image retrieval) and low-level (image inpainting) tasks. Our OCNNs also generate realistic images (Section \ref{sec:gen}) and are more robust to adversarial attacks (Section \ref{sec:attack}). 

\subsection{Network Dissection }
We demonstrate in Section \ref{sec:exp} that OCNNs reduce the feature redundancy by decorrelating different feature channels and enhancing the feature expressiveness with improved performance in image retrieval, inpainting, and generation. Network dissection \cite{bau2017network} is utilized to further evaluate the feature expressiveness across different channels.

Network dissection \cite{bau2017network} is a framework that quantifies the interpretability of latent representations of CNNs by evaluating the alignment between individual hidden units and a set of semantic concepts. Specifically, we evaluate the baseline and our OCNN with backbone architecture ResNet34 \cite{he2016deep} trained on ImageNet. Models are evaluated on the Broden \cite{bau2017network} dataset, where each image is annotated with spatial regions of different concepts, including cat, dog, house, etc. The concepts are further grouped into 6 categories: scene, object, part, material, texture, and color. Network dissection framework compares the mean intersection over union (mIoU) between network channel-wise activation of each layer and ground-truth annotations. The units/feature channels are considered as ``effective'' when mIoU$\geq 0.04$. They are denoted as ``unique detectors''.

\begin{figure}[!t]
\centering
        \includegraphics[width=0.4\textwidth]{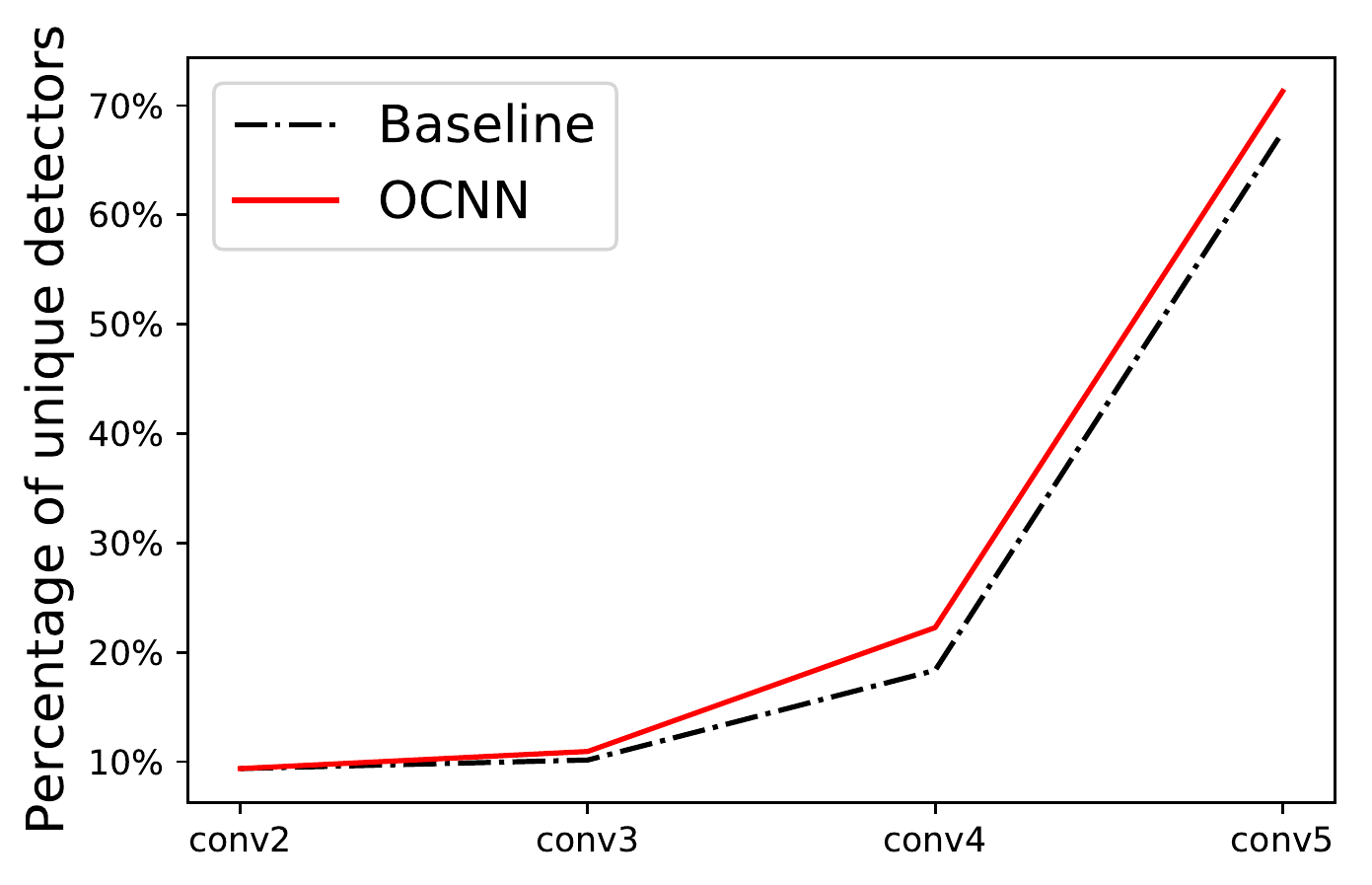}%
       \caption{\small  Percentage of unique detectors (mIoU $\geq$ 0.04) over different layers. Our OCNN has more unique detectors compared to plain baseline ResNet34 \cite{he2016deep} at each layer.}
\label{fig:comp_dise}
\end{figure}

\begin{figure}[!t]
        \includegraphics[width=0.5\columnwidth]{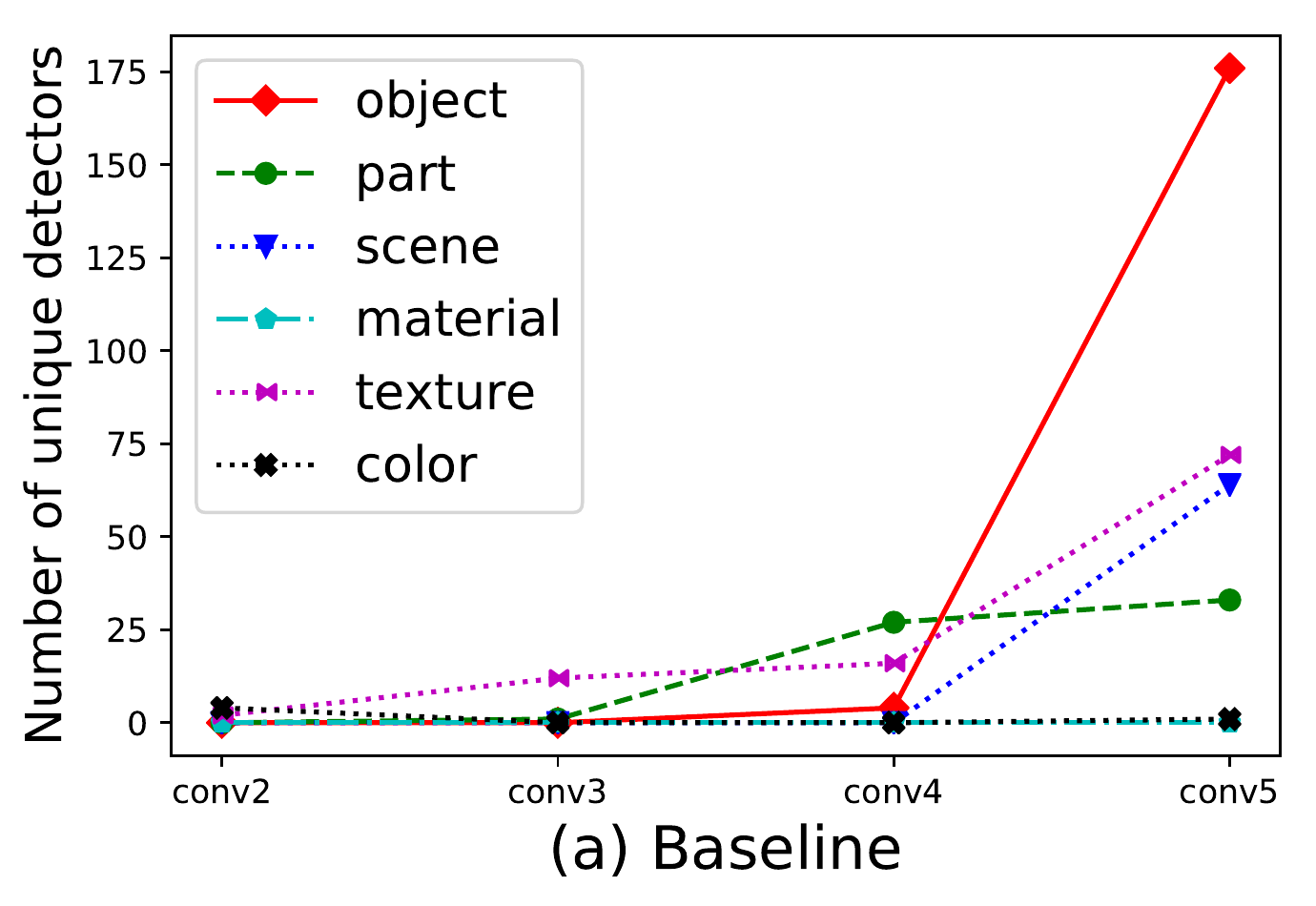}%
        \includegraphics[width=0.5\columnwidth]{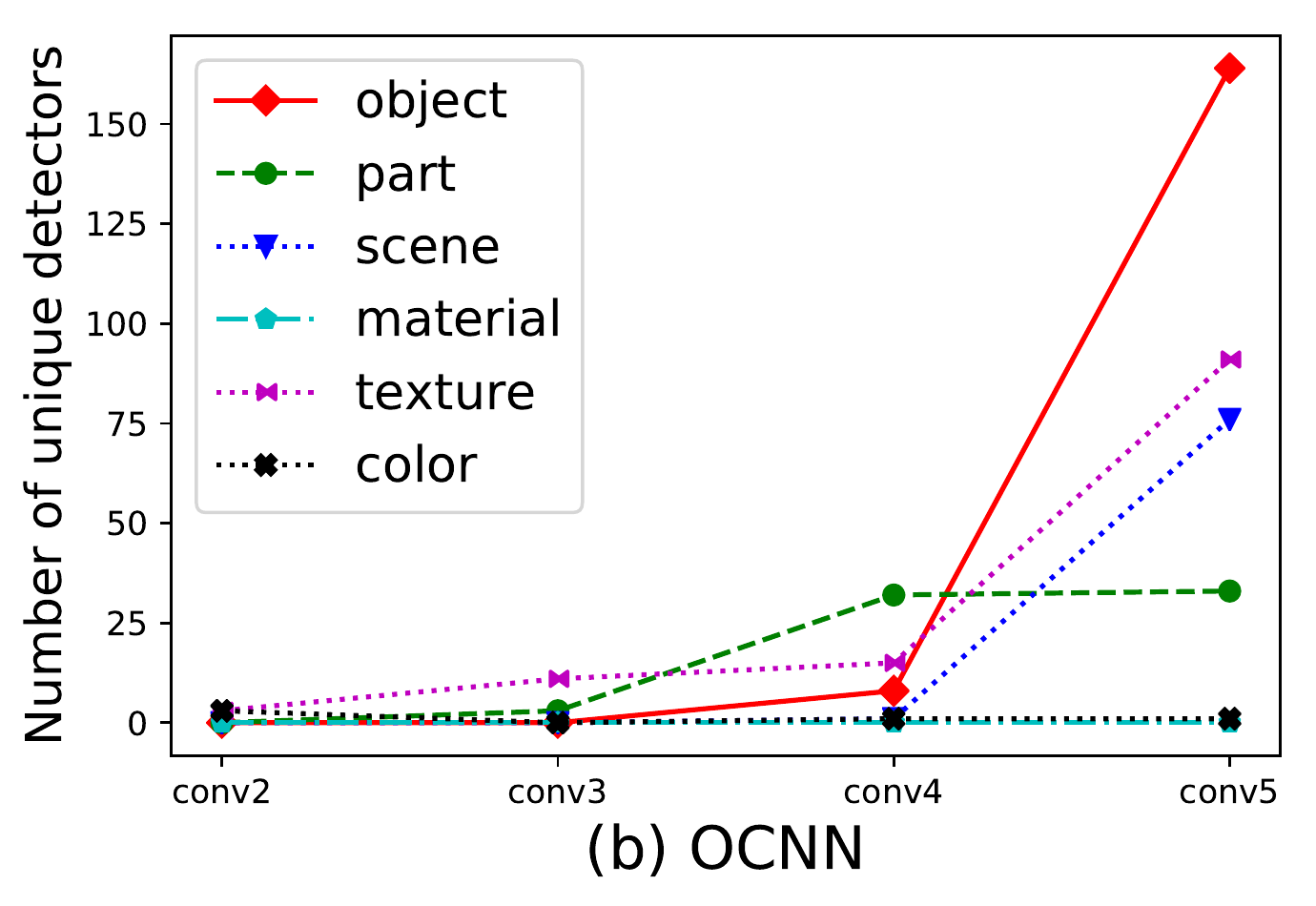}%
       \caption{\small Distribution of concepts of unique detectors of different layers. Our OCNN has more uniform concept distribution compared to plain baseline ResNet34 \cite{he2016deep}.}
\label{fig:dist_dise}
\end{figure}

 OCNNs  have more unique detectors over different layers of the network (Table \ref{tab:dise} \& Fig.\ref{fig:comp_dise}). Additionally, OCNNs have more uniform distributions of 6  concept categories (Fig.\ref{fig:dist_dise}). These imply that orthogonal convolutions reduce feature redundancy and enhance the feature expressiveness.

\begin{table}[!ht]
\caption{Number of unique detectors (feature channels with mIoU $\geq$ 0.04) comparisons on ImageNet \cite{deng2009imagenet}.}
\begin{center}
\resizebox{0.4\textwidth}{!}{
\begin{tabular}{>{\columncolor[gray]{0.95}}c|c|c|c|c}
\hline
\rowcolor{LightCyan}     & conv2 & conv3 & conv4 & conv5 \\ \hline
ResNet34 \cite{he2016deep} & 6     & 13    & 47    & 346   \\ \hline
OCNN  (ours)    & 6     & 14    & 57    & 365   \\ \hline
\end{tabular}
}
 \vspace{-1.5em}

\end{center}

\label{tab:dise}
\end{table}

\label{sec:disec}
   
\subsection{Deep Metric Learning}
\label{sec:DML}
We evaluate the generalizability and performance of our our orthogonal regularizer in deep metric learning tasks. Specifically, following the training/evaluation settings in \cite{movshovitz2017no}, we perform retrieval and clustering on Cars196 dataset \cite{KrauseStarkDengFei-Fei_3DRR2013} and summarize the results in Table \ref{tab:cars}. We observe performance gains when the orthogonal regularizer is added.

\begin{table}[!ht]

\begin{center}
\centering
\caption{\small  Retrieval/clustering performance on Cars196 (\%).}
\label{tab:cars}
\resizebox{\columnwidth}{!}{
\begin{tabular}{>{\columncolor[gray]{0.95}}c|c|c|cccc}
\hline
\rowcolor{LightCyan}          & NMI    & F1     & Recall@1 & @2     & @4     & @8     \\ \hline
Triplet loss \cite{hoffer2015deep}     & 61.9 & 27.1 & 61.4   & 73.5 & 83.1 & 89.9 \\ \hline
ProxyNCA \cite{movshovitz2017no}    & 62.4 & 29.2 & 67.9   & 78.2 & 85.6 & 90.6 \\ \hline
\cite{movshovitz2017no}+Kernel orth & 63.1 & 29.6 & 67.6   & 78.4 & 86.2 & 91.2 \\ \hline
\cite{movshovitz2017no}+OCNN        & \textbf{63.6} & \textbf{30.2} & \textbf{68.8}   & \textbf{79.0} & \textbf{87.4} & \textbf{92.0} \\ \hline
\end{tabular}
}
\end{center}
\vspace{-0.25in}

\end{table}

\subsection{Proof of the Orthogonality Equivalence}
\label{sec:proof}
Here we provide a proof for the lemma~\ref{lemma:equivalence}: The row orthogonality and column orthogonality are equivalent in the MSE sense, i.e. $\|\mathcal{K}\mathcal{K}^T - I\|_F^2 = \|\mathcal{K}^T\mathcal{K} - I^\prime\|_F^2 + U$, where $U$ is a constant.
A simple motivation for this proof is that when $\mathcal{K}$ is a square matrix, then $\mathcal{K}\mathcal{K}^T = I \iff \mathcal{K}^T\mathcal{K} = I^{\prime}$. So we can hope to generalize this result and provide a more convenient algorithm. The following short proof is provided in the supplementary material of \cite{le2011ica}. We would like to present it here for the reader's convenience.

\begin{proof}
It's sufficient to prove the general result, where we choose $\mathcal{K}\in \mathbf{R}^{M\times N}$ to be an arbitrary matrix\footnote{Here $M$ and $N$ are just some constant, different from the the ones used in the main text.}. We denote $\|\mathcal{K}\mathcal{K}^T - I_M\|^2$ as $L_{r}$ and $\|\mathcal{K}^T\mathcal{K} - I_N\|^2$ as $L_c$.
\begin{align*}
L_r &= \|\mathcal{K}\mathcal{K}^T - I_M\|_F^2 \\
&= tr\left[(\mathcal{K}\mathcal{K}^T - I_M)^T(\mathcal{K}\mathcal{K}^T - I_M)\right]\\
&= tr(\mathcal{K}\mathcal{K}^T\mathcal{K}\mathcal{K}^T) - 2tr(\mathcal{K}\mathcal{K}^T) + tr(I_M)\\
&= tr(\mathcal{K}\mathcal{K}\mathcal{K}^T\mathcal{K}) - 2tr(\mathcal{K}^T\mathcal{K}) + tr(I_N) + M - N\\
&= tr\left[\mathcal{K}^T\mathcal{K}\mathcal{K}^T\mathcal{K} - 2\mathcal{K}^T\mathcal{K} + I_N\right] + M - N\\
&= tr\left[(\mathcal{K}^T\mathcal{K} - I_N)(\mathcal{K}^T\mathcal{K} - I_N)\right] + M - N\\
&= \|\mathcal{K}^T\mathcal{K} - I_N\|_F^2 + M - N\\
&= L_c + U
\end{align*}
where $U = M - N$.

\end{proof}

\subsection{Filter Similarity Visualizations}
\label{sec:feat_sim}
As shown in Fig.\ref{fig:intro}, filter similarity increases with depth of the network. We visualize the guided back-propagation patterns to understand this phenomenon. 

\begin{figure}[t!]
   \vspace{-0.5em}

\centering
        \includegraphics[width=\columnwidth]{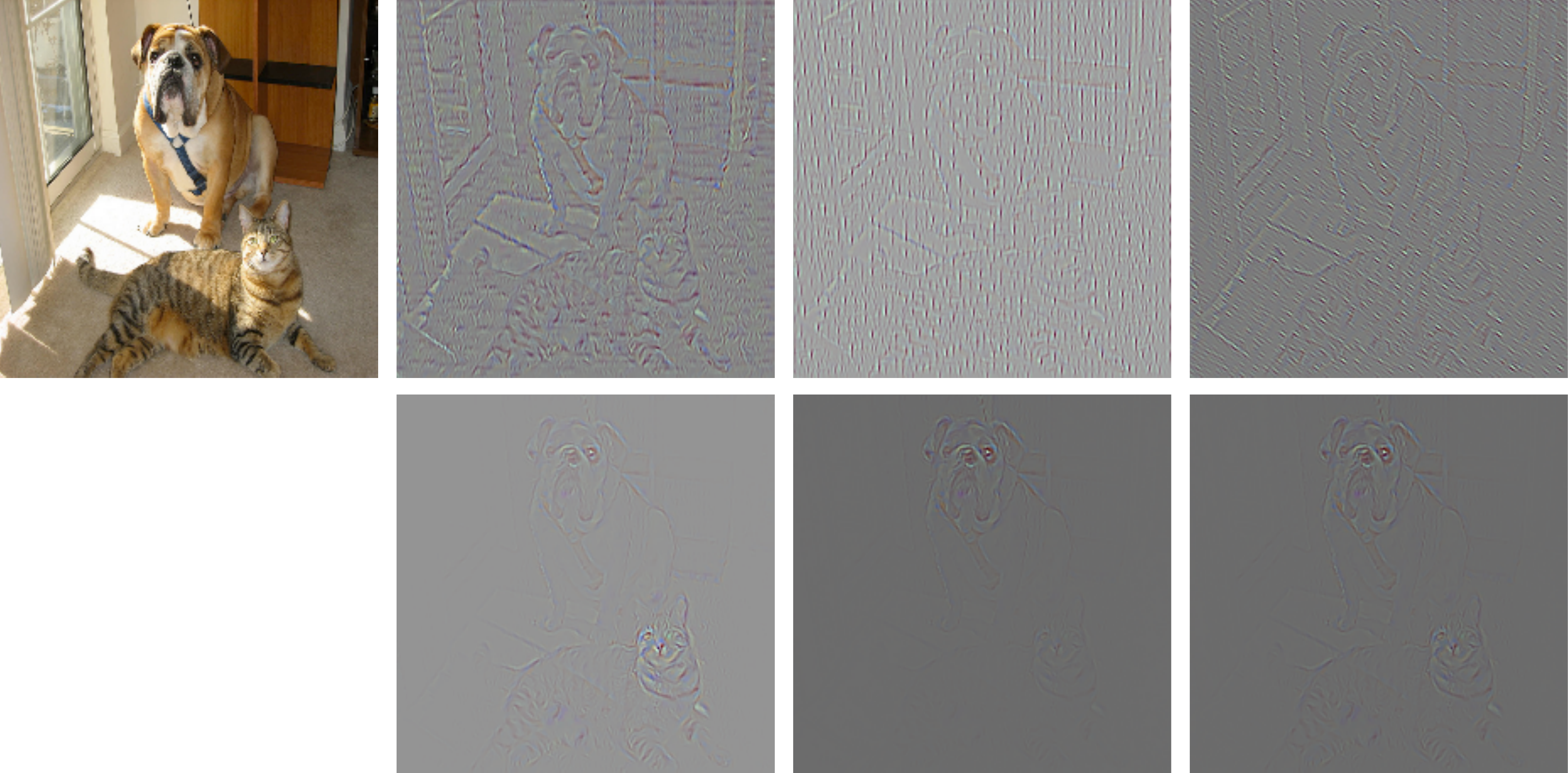}%
   \caption{\small Guided back-propagation patterns of the input image (first column) with a ResNet34 model. The first row depicts patterns of the first 3 channels from layer 7, while the second row depicts patterns of the first 3 channels from layer 33. The filter similarity increases with network depth.}
   \vspace{-1em}
   \label{fig:gbp}
   \end{figure}

For the ResNet34 trained on ImageNet, we plot guided back-propagation patterns of an image in Fig.\ref{fig:gbp}. The first row depicts patterns of the first 3 channels from layer 7, while the second row depicts patterns of the first 3 channels from layer 33. Patterns of different channels from earlier layers are more diverse, while patterns of different channels from later layers usually focus on certain regions. The filter similarity increases with depth.

